\documentclass[11pt]{article}

\usepackage{EMNLP2023}
\usepackage{times}
\usepackage{latexsym}
\usepackage[T1]{fontenc}
\usepackage[utf8]{inputenc}
\usepackage{microtype}
\usepackage{inconsolata}
\usepackage{enumitem}
\usepackage{graphicx}
\usepackage{svg}
\usepackage{booktabs}
\usepackage{amsfonts}
\usepackage{amsmath}
\usepackage{makecell}
\usepackage{array}
\usepackage{multirow}
\usepackage{xspace}
\usepackage{xcolor}
\usepackage{url}
\usepackage{float} 
\usepackage{amsmath}
\usepackage[most]{tcolorbox} 
\usepackage{tcolorbox}

\newcommand{\modelname}{TranslatePsy-AfriSLM\xspace}

\title{\modelname: High-Quality Data Scaling \\
For Low-Resource Machine Translation}

\author{
  Milan Gritta\thanks{\ \ Equal contribution.} \and
  Patrik Lambert\footnotemark[1] \and
  Jihye Back\footnotemark[1] \and
  Amril Nazir \\
  Tether AI Research \\
  \texttt{\{milan.gritta, patrik.lambert, jihye.back, amril.nazir\}@tether.io}
}
\begin{document}
\maketitle

\begin{abstract}

The rapid progress in Artificial Intelligence has largely bypassed African languages, creating a digital divide that limits AI adoption on the continent. Recent open-source LLMs systematically underperform on African language machine translation, while the lack of large-scale, high-quality, open-source parallel data has constrained the development of competitive small language models (SLMs). We introduce \textbf{\modelname}, a collection of open-source machine translation resources for 19 Sub-Saharan African languages, including curated parallel data, African-specialized synthetic data, and a family of fine-tuned SLMs. Our empirical study shows that unified quality-estimation filtering removes up to 96\% of training tokens without degrading quality, and that filtered synthetic data dominates the quality-efficiency Pareto frontier. Fine-tuned on the resulting data mixture, TranslatePsy-AfriSLMs outperform substantially larger systems, including TranslateGemma-27B and Qwen3.5-122B-A10B, with as few as 0.8B parameters.

\end{abstract}

\section{Introduction}
\label{sec:intro}

\begin{figure}[t]
    \centering
    \includegraphics[width=0.99\linewidth]{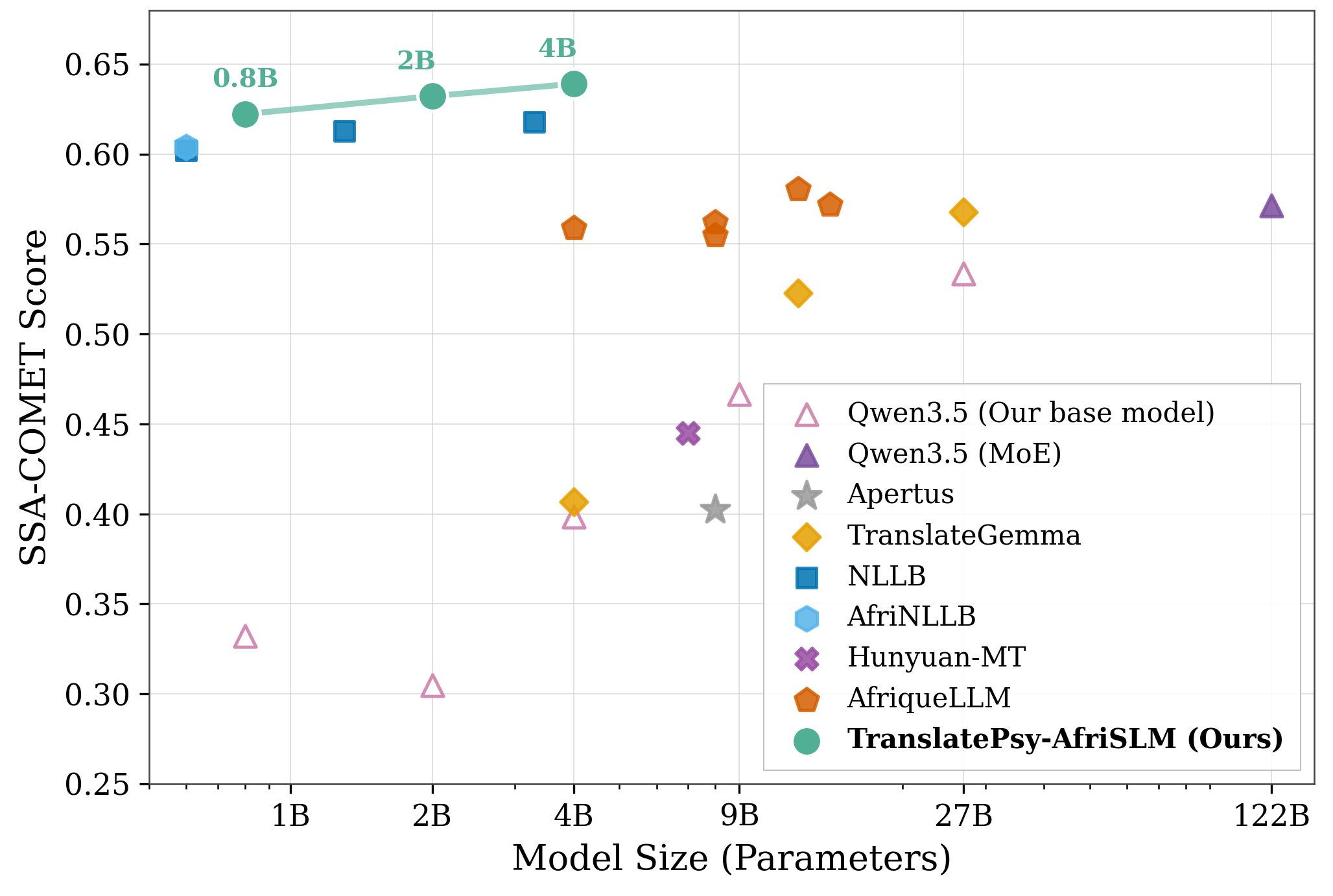}
    \caption{\textbf{TranslatePsy-AfriSLMs outperform frontier LLMs} such as TranslateGemma-27B. Figure shows SSA-COMET scores on the BOUQuET benchmark. 
    }
    \label{fig:bouquet_scaling}
\end{figure}

The AI underinvestment on the African continent \cite{nwagbala2025ai,diallo2025case,isangula2025navigating} has created a significant adoption barrier for over a billion people, preventing them from fully exploiting the productivity and collaboration benefits that AI can offer \cite{maluleke2025ai}. A prime example of this is a lack of performant SLMs for Machine Translation (MT), a critical utility for facilitating cross-border communication, trade, and education \cite{ssemugabi2025role,moukatib2026role}. However, most frontier open-source LLMs such as Apertus \cite{swissai2025apertus}, Qwen3 \cite{yang2025qwen3}, TranslateGemma \cite{finkelstein2026translategemma}, Hunyuan-MT \cite{zheng2025hunyuan} or Qwen3.5 \cite{qwen35blog} systematically underperform on African language MT, see Figure \ref{fig:bouquet_scaling}, while incurring substantial running costs due to large parameter counts.

\begin{figure*}[t]
\centering
\includegraphics[width=\linewidth]{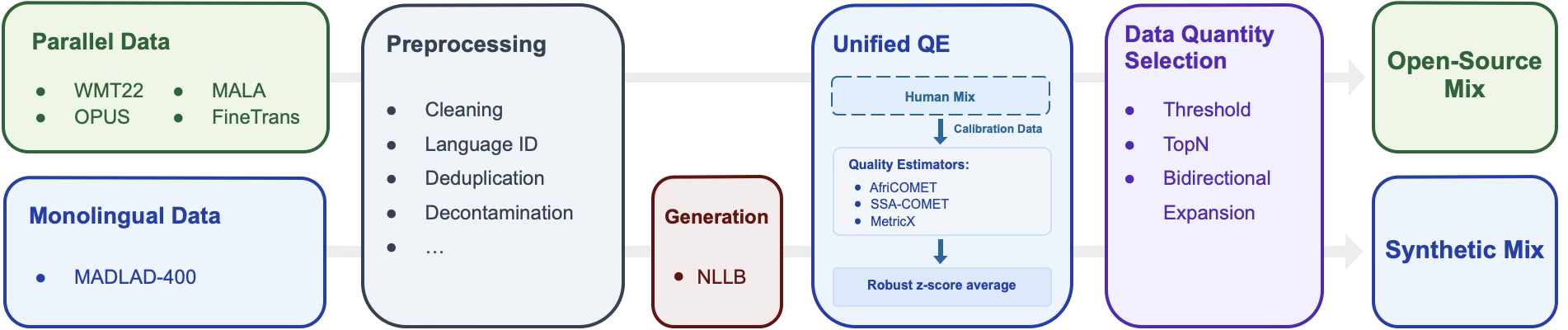}
\caption{\textbf{\modelname data pipeline.} Parallel and monolingual sources are unified through preprocessing, QE filtering, calibrated on the Human Mix. Finally, the optimal Open-Source and Synthetic mixes are selected.}
\label{fig:data_pipeline}
\end{figure*}

Furthermore, efficiently adapting existing SLMs to support this task remains challenging due to the scarcity of high-quality, open-source data in sufficient quantities, as our survey shows in Background (\S\ref{sec:background}). Large internet repositories are highly unstructured and noisy, the training signal is sparse and the token budget dramatically inflated. Existing curated datasets and human-quality translations are too small, offering only modest improvements. 
Therefore, to support efficient model adaptation for low-resource MT, we introduce \textbf{\modelname}\footnote{ \url{https://huggingface.co/collections/qvac/translatepsy-afrislm}}, an open-source African language MT resource comprising high-quality parallel data for 19 Sub-Saharan African languages and a family of highly-capable SLMs (Table \ref{tab:ssa_results_short} and Figure \ref{fig:bouquet_scaling}).
Our approach is systematically validated on key benchmarks, with deep insights and analyses that can guide future research. \modelname models exceed African language translation skills of frontier LLMs such as TranslateGemma-27B and Qwen3.5-27B (and 122B-A10B) with only 0.8B parameters. Our smallest model even outperforms the dedicated NLLB encoder-decoder models while preserving conversational capabilities (Figure \ref{fig:multilingual_mt_conversation}).

\section{Background}
\label{sec:background}

\subsection{African language MT Training Resources}

Existing African MT datasets reveal a persistent trade-off between scale, coverage, and translation quality. We broadly group them into 3 categories: large but unstructured repositories, medium-sized curated corpora, and small, human-quality datasets.

\paragraph{Large but Unstructured.} 
Data repositories such as OPUS \cite{tiedemann-2012-parallel}, MALA \cite{ji2024emma500enhancingmassivelymultilingual}, WMT22 \cite{adelani-etal-2022-findings} and Fine Translations \cite{penedo2026finetranslations} contain large quantities of parallel data. However, they vary widely in volume, language coverage, and translation direction, while also suffering from high duplication rates, test set contamination, and noisy text. NLLB \cite{team2022language} leveraged approximately 18 billion sentence pairs across 200 languages, but did not release a fixed, high-quality corpus suitable for the token-efficient adaptation of African language MT models.

\paragraph{Medium-Sized and Curated.} AfriNLLB\footnote{\scriptsize\url{https://huggingface.co/datasets/AfriNLP/AfriNLLB-train}} is currently the only readily available, medium-sized, curated dataset for MT \cite{Moslem2026AfriNLLBET}. However, it covers only 9 African languages, and approximately 50\% of its $\sim$3 million pairs include Arabic or European languages. AfriqueLLM \cite{yu2026afriquellm} adapts frontier LLMs for African languages through continued pretraining. While its general-purpose scope significantly improves MT, its training data was not publicly released (a gap we aim to fill). We report comparisons with AfriqueLLM models in Figure~\ref{fig:bouquet_scaling} and SLMs trained on AfriNLLB data in Section \ref{sec:our_final_models}.

\paragraph{Small but Human-Quality.}

Several datasets are available in this subset, e.g. MMT-Africa \cite{emezue-dossou-2021-mmtafrica}, AfriDOC-MT \cite{alabi2025afridoc}, SMOL \cite{caswell2025smol}, LAFAND-MT \cite{adelani-etal-2022-thousand}, WMT24pp \cite{deutsch2025wmt24expandinglanguagecoverage}, MENYO-MT \cite{adelani-etal-2021-effect}. However, they occur in limited quantities and have uneven language coverage. We show that despite their high-quality translations, such mixtures yield only modest improvements (\S\ref{sec:data_source_results}), smaller than AfriNLLB and substantially smaller than our best \modelname mixes. This further motivates our large-scale, quality-focused data curation.

\subsection{Quality Estimation for Data Filtering}
\label{sec:qe_for_filtering}

Afri-COMET~\cite{wang2024afrimte} and more recently, SSA-COMET~\cite{li-etal-2025-ssa} were introduced as African-centric, reference-free alternatives to COMET~\cite{rei-etal-2020-comet}, COMET-KIWI~\cite{rei-etal-2022-cometkiwi} and MetricX~\cite{juraska-etal-2023-metricx}. Prior works have used them to filter parallel text in African languages~\cite{yu2026afriquellm,uemura-etal-2026-afrimteb,Moslem2026AfriNLLBET}.
However, two questions remain unanswered: (1) the comparative usefulness of each metric \emph{as a training data filter}, as opposed to an evaluation metric, validated on large-scale African language MT; and (2) whether a combination of individual metrics would perform more consistently, and if so, how to effectively combine multiple QE metrics (\S\ref{sec:afri_data}).

\section{\modelname}
\label{sec:afri_data}

We address the heterogeneity of African language MT data with a multi-step curation pipeline (Figure~\ref{fig:data_pipeline}) applied to open-source and synthetic sentence pairs (\S\ref{sec:data_sources}). After standard structural preprocessing, we score raw pairs using Unified Quality Estimation (\S\ref{sec:quality_estimation}), a robust z-score that aggregates multiple QE metrics. We then study data quantity selection methods for post-training (\S\ref{sec:quantity}).

\subsection{Data Sources}
\label{sec:data_sources}

\paragraph{Parallel Data.}
\label{sec:filtered_data}

We source data from four large, open-source repositories: WMT22 \cite{adelani-etal-2022-findings}, MALA \cite{ji2024emma500enhancingmassivelymultilingual}, OPUS \cite{tiedemann-2012-parallel} and Fine Translations \cite{penedo2026finetranslations}. Unlike synthetic data, where we can control generation volume and direction, open-source corpora have fixed and often highly uneven distributions across languages and translation directions. As a result, these variables can \textbf{vary significantly} between sources. Therefore, we cap each language pair to a maximum of 5 million sentence pairs to mitigate 'winner-takes-all' effects. This process gives us approximately 427 million raw sentence pairs, see Table~\ref{tab:open_source_mix} for a per-language breakdown.

\paragraph{Monolingual Data.}
\label{sec:synthetic_data}
We used the MADLAD-400 corpus~\cite{kudugunta2023madlad400} to source synthetic data, see "Synthetic Data Generation" below. For each of the 19 African languages considered in this work, we processed all available data. For English, due to the vast volume of data available, we randomly sampled 3.6 million documents. We report detailed, per-language statistics in Table~\ref{tab:synthetic_mix}.

\paragraph{Preprocessing.} This includes standard cleaning, language checks, deduplication, and test set decontamination. Details are provided in Appendix~\ref{appendix:data_processing}.

\paragraph{Synthetic Data Generation}
To generate synthetic data, we translated the monolingual data with the NLLB-3.3B encoder-decoder, selected as the teacher model based on our benchmarks in Table~\ref{tab:ssa_results}. See Appendix~\ref{appendix:data_processing} for generation settings.

\subsection{Unified Quality Estimation}
\label{sec:quality_estimation}
We aim to filter the open-source and synthetic data with AfriCOMET,\footnote{\scriptsize\url{https://huggingface.co/Masakhane/africomet-qe-stl-1.1}} SSA-COMET\footnote{\scriptsize\url{https://huggingface.co/McGill-NLP/ssa-comet-mtl} - according to \citet{li-etal-2025-ssa}, this model achieves better results than \textbf{ssa-comet-qe}.} and MetricX\footnote{\scriptsize\url{https://huggingface.co/google/metricx-24-hybrid-xl-v2p6 without reference.}} QE metrics\footnote{\scriptsize\textbf{Note:} These metrics are used for both quality estimation \textbf{and} reference-based evaluation. We use the names interchangeably depending on local context.}.
As we show later (§\ref{sec:data_estimators_results}), no single quality estimator consistently performs best across all evaluation metrics. This motivates aggregating all three estimators into a single score for more consistent filtering. However, a naïve aggregation seems impractical, as the scores have distinct polarities (higher- versus lower-is-better) and their ranges vary across QE models, language directions and training corpora. To address this, we unify the quality estimates from AfriCOMET, SSA-COMET and MetricX by mapping them into a shared \textbf{robust z-score} (Eq.~\ref{eq:z}), calibrated on the following dataset. 

\paragraph{Human Mix.} 

We compiled \textasciitilde352~k high-quality, human-translated pairs from two datasets (see Appendix~\ref{human} for details) for two distinct purposes: a) to provide a human-quality reference for the robust z-score parameters, and b) to show that high-quality but limited-scale human-translated data alone is insufficient for effective post-training adaptation.

\paragraph{Average Robust z-score.}
For each translation direction $d$ and for each metric $m$, we compute Human Mix calibration statistics: the median $\mathrm{\tilde{x}_{d,m}}$ and the Median Absolute Deviation (MAD):
\begin{equation}
\label{eq:mad} \mathrm{\operatorname{MAD}_{d,m} = \operatorname{median}\!\left( \left| x_i - \tilde{x}_{d,m} \right| \right).} \end{equation}
Each candidate QE score $\mathrm{x_{i,d,m}}$ is then normalized against these statistics:
\begin{equation}
\label{eq:z}
    \mathrm{z_{i,d,m} = s_m \cdot \frac{0.6745\,(x_{i,d,m} - \tilde{x}_{d,m})}{\operatorname{MAD}_{d,m}}}
\end{equation}
where $\mathrm{s_m \in \{+1, -1\}}$ flips lower-is-better metrics such as MetricX, so that higher values always indicate higher quality.
By calibrating each score against the same Human Mix statistics (Eq. \ref{eq:z}), this normalization places heterogeneous data sources, translation directions, and QE metrics on a common quality scale relative to human-translated data. They can now be aggregated by computing the average of the per-metric normalized scores in Equation \ref{eq:avg_z-score}. 
We adopt this \textit{average robust z-score} as a unified QE metric, henceforth denoted simply as $\mathrm{\bar{z}}$:
\begin{equation}
\label{eq:avg_z-score}
    \mathrm{\bar{z}_{i,d} = \frac{1}{3}\sum_{m=1}^{3} z_{i,d,m}}
\end{equation}
This unified score provides a more consistent basis for filtering heterogeneous parallel data.

\paragraph{Directional Filtering.}
Quality estimation scores are not symmetric with respect to their \textbf{argument order}; consequently, for any training pair $\mathrm{(X, Y)}$ the score $\mathrm{\bar{z}(X, Y)}$ typically differs from $\mathrm{\bar{z}(Y, X)}$. The choice of scoring direction therefore determines which sentence pairs pass the filter. We investigate the following strategies for filtering:

\begin{samepage}
\begin{itemize}[itemsep=0mm]
    \item \textbf{Aligned}: $\mathrm{\bar{z}(X, Y)}$, train $\mathrm{X \rightarrow Y}$
    \item \textbf{Reversed}: $\mathrm{\bar{z}(Y, X)}$, train $\mathrm{X \rightarrow Y}$
    \item \textbf{Mean}: $\mathrm{\tfrac{1}{2}\bigl(\bar{z}(X, Y) + \bar{z}(Y, X)\bigr)}$, train $\mathrm{X \rightarrow Y}$
\end{itemize}
\end{samepage}

\noindent The \textit{aligned} strategy scores each pair in the training direction, whereas the \textit{reversed} strategy scores it in the opposite direction. The \textit{mean} strategy averages the z-scores from both directions, potentially providing a balanced alternative that we also explore.

%For each strategy, pairs for which the corresponding z-score is above a global threshold are retained. 

\subsection{Data Quantity Selection} 
\label{sec:quantity}
Once the QE strategy is determined, we need to investigate methods of selecting training data quantities: 1) \textbf{Threshold} filters out sentence pairs below a given $\bar{z}$ score, 2) \textbf{TopN} extends the threshold filter by limiting any single language pair to a maximum of top-N examples to mitigate large imbalances between languages, 3) \textbf{Bidirectional expansion} is an orthogonal data augmentation step that reverses each sentence pair $\mathrm{\mathrm{(X, Y) \rightarrow (Y, X)}}$ to balance translation directions during training.

\subsection{Auxiliary Data Mixes}
\label{sec:additional_mixes}

In addition to the open-source and synthetic mixes, we include two auxiliary mixtures to maintain capabilities related to broader machine translation. 

\paragraph{Instruct Mix} is our multilingual instruction-following dataset with approximately 50\% African-language content, totalling 4.6 million examples. We include it to preserve general conversational abilities and to extend translation to multi-turn settings. The dataset composition is provided in \S\ref{sec:instruction_mix_appendix}.

\paragraph{Asia-Europe Mix} covers 38 languages ($\sim$24 million examples). We include it to preserve translation quality on medium- and high-resource languages and to study whether such data mitigates catastrophic forgetting (dataset details in \S\ref{sec:europe_asia_appendix}).

\section{Experimental Setup}
\label{sec:experimental_setup}

\subsection{Supervised Fine-Tuning (SFT)}
\label{sec:sft}

We post-train Qwen3.5~\cite{qwen35blog} models using Supervised Fine-Tuning (SFT), as they consistently outperform other general-purpose SLMs in our preliminary benchmarks.\footnote{We have omitted larger African-specialized models like AfriqueLLM as backbones because \modelname targets high-performing SLMs deployable on low-cost devices. AfriqueLLMs start at over 4 billion parameters, hence the inference cost is likely prohibitive for our intended users.} Each model undergoes (full-parameter) fine-tuning for one epoch on the final training mixture selected from our data analyses in \S\ref{sec:pareto_frontier}, with loss computed only on assistant tokens. Details are provided in Appendix~\ref{app:hyperparams}.

\begin{figure*}[t]
\centering
\includegraphics[width=\linewidth]{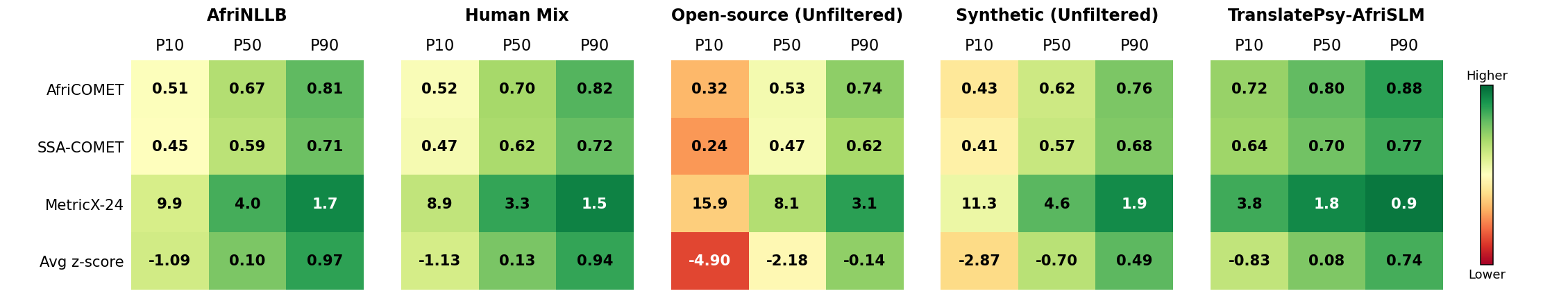}
\caption{\textbf{Comparison of data quality across data sources.} Each column shows the P10, P50, and P90 percentiles for AfriCOMET, SSA-COMET and MetricX, which are then normalised into a mean robust z-score (bottom row).}
\vspace{-0.5em}
\label{fig:qe_heatmap}
\end{figure*}

\subsection{Evaluation}
\label{sec:evaluation}

\paragraph{Language groups.} 
% iid
% ood
We evaluate MT on 19 languages covered in the \modelname data mixes, labelled \textbf{Africa-IID}\footnote{\scriptsize Africa-IID: Afrikaans (afr), Amharic (amh), Hausa (hau), Igbo (ibo), Kinyarwanda (kin), Lingala (lin), Luganda (lug), Malagasy (plt/mlg), Nyanja (nya), Oromo (orm), Nigerian Pidgin (pcm), Shona (sna), Somali (som), Southern Sotho (sot), Swahili (swh/swa), Tswana (tsn), Wolof (wol), Xhosa (xho), Yoruba (yor), Zulu (zul); see \url{https://en.wikipedia.org/wiki/ISO_639-3}.}, and an additional set of 8 unseen languages, labelled \textbf{Africa-OOD}\footnote{\scriptsize Africa-OOD: Nigerian Pidgin (pcm), Sudanese Arabic (apd), Akan (aka), Tamazight (ber), Kituba (ktu), Bambara (bam), Sepedi (nso) and Mooré (mos).}, to assess generalisation to out-of-domain languages. 

\paragraph{Benchmarks.} We evaluate on three translation benchmarks: Flores-200 \cite{team2022language}, BOUQuET \cite{omnilingual2026}, and Smol \cite{caswell2025smol}. Flores-200 is the most widely used benchmark (\texttt{devtest} split, 1,012 sentences), enabling comparison with prior work. BOUQuET is the most recent, linguist-curated multi-way benchmark (\texttt{test} split, 854 sentences) with broad typological coverage. Smol (\texttt{smolsent} split, 863 sentences) provides professionally translated data across 89 languages, used exclusively for evaluation. Each benchmark covers all 19 Africa-IID languages.

%reorder -> Comet-22 SSA MetricX chrF
\paragraph{Metrics.} We evaluate translation quality using five complementary metrics. We report three learned metrics: \textbf{COMET-22}\footnote{\scriptsize\url{https://huggingface.co/Unbabel/wmt22-comet-da}} \cite{rei-etal-2022-comet}, a neural metric trained on human direct assessments, abbreviated to \textbf{C22}; \textbf{SSA-COMET}\footnote{\scriptsize\url{https://huggingface.co/McGill-NLP/ssa-comet-mtl}} \cite{li-etal-2025-ssa}, an AfroXLM-R-based COMET variant developed for evaluating African languages, abbreviated to \textbf{SSA}; and \textbf{MetricX}\footnote{\scriptsize\url{https://huggingface.co/google/metricx-24-hybrid-xl-v2p6}} \cite{juraska-etal-2024-metricx}, an mT5-based regression metric fine-tuned on DA and MQM ratings, abbreviated to \textbf{MX}. We additionally report two reference-based metrics that are not used for QE-based filtering: \textbf{chrF++}\footnote{\scriptsize SacreBLEU with character order 6, word order 2, beta 2, case-sensitive evaluation, whitespace excluded, and epsilon smoothing disabled} \cite{popovic-2017-chrf}, a character- and word-based n-gram F-score, and \textbf{spBLEU}\footnote{\scriptsize SacreBLEU using the Flores-200 tokenizer and its SentencePiece model} \cite{goyal2021flores1101}, BLEU computed over SentencePiece-tokenized text using the Flores-200 tokenizer. 
COMET-22 and SSA-COMET are reported on a 0--1 scale (higher is better), chrF++ and spBLEU on a 0--100 scale (higher is better), and MetricX on a 0--25 scale (lower is better). Since some of the learned metrics used for evaluation overlap with the QE estimators used for data selection, chrF++ and spBLEU provide complementary signals independent of the filtering process. This allows us to assess whether the observed improvements generalize beyond the metrics involved in QE-based filtering.

\paragraph{Baselines} We benchmark our SLMs against \textbf{(1) General-purpose LLMs:} Qwen3~\cite{yang2025qwen3} and Qwen3.5~\cite{qwen35blog}, two strong general-purpose multilingual LLM families, and Apertus~\cite{swissai2025apertus}, a massively multilingual model covering 1,800+ languages. Among these, Qwen3.5 achieves the strongest performance, motivating its use as our backbone and enabling a direct measurement of gains from our methodology. \textbf{(2) Dedicated translation models:} NLLB~\cite{team2022language}, an encoder--decoder model trained on large-scale parallel data; AfriNLLB~\cite{Moslem2026AfriNLLBET}, an African-centric variant of NLLB; TranslateGemma~\cite{finkelstein2026translategemma} and Hunyuan-MT~\cite{zheng2025hunyuan}, decoder-only LLMs post-trained for translation. \textbf{(3) African language-specialized models:} AfriqueLLM~\cite{yu2026afriquellm}, which adapts Gemma-3, Qwen3, and LLaMA-3.1 via continued pretraining on approximately 26B tokens.
% Table~\ref{tab:ssa_results} groups models by these categories. 

\section{Results and Analysis}
\label{sec:results}

\subsection{Quality Estimation: Which is Best?}
\label{sec:data_estimators_results}

We first study which quality estimation strategy is most effective for data filtering. All experiments in this section use synthetic data from English into African languages, the only corpus large enough to support controlled comparisons across all 19 language pairs, with over 135 million examples.

\paragraph{No Single Estimator Is Optimal.}

We compare SSA-COMET, AfriCOMET, and MetricX as individual quality estimators for filtering data. For each estimator, we score the full training pool%in both translation directions
, select the top 2 million examples per language, and use them to fine-tune Qwen3.5-2B. 
Table~\ref{tab:indiv_qe_metrics_gain} shows the relative performance gains over a randomly selected baseline, averaged across BOUQuET, Flores-200, and Smol test sets.\footnote{In the appendix, Table~\ref{tab:indiv_qe_metrics} shows the metrics scores, and Table~\ref{tab:qe_metrics_significance} paired bootstrap significance tests.}
%Flores-200 and Smol follow the same pattern, as shown in Tables~\ref{tab:indiv_qe_metrics_flores} and ~\ref{tab:indiv_qe_metrics_smol}. 
First, filtered data yield better translation scores than randomly selected data in nearly all cases. For example, even data selected using MetricX achieve a higher SSA-COMET score than the baseline. However, the overall impact of filtering is moderate, suggesting that the initial quality of the synthetic training pool was already strong. 
Second, each estimator performs best when evaluated by its corresponding metric. For instance, the SSA-COMET estimator consistently achieves the top SSA-COMET score, and the same holds for MetricX. Similarly, the AfriCOMET estimator outperforms MetricX QE when evaluated via COMET variants. Consequently, no single estimator dominates across all evaluation metrics. However, the $\bar{z}$ score does provide a balanced alternative, achieving \textbf{at least the second best score across every metric}. We therefore adopt it as our QE metric for the remainder of the paper.

\begin{table}[h]
\centering
\small
\resizebox{\columnwidth}{!}{
\begin{tabular}{lcccc}
\toprule
Estimator & C22 & SSA & MX & chrF++ \\
\midrule
Random & 0 & 0 & 0 & 0 \\
z-score & \textbf{+1.08\%} & +4.32\% & +9.57\% & +0.73\% \\
SSA-COMET & +0.94\% & \textbf{+4.85\%} & +5.84\% & \textbf{+1.22\%} \\
AfriCOMET & +0.80\% & +3.02\% & +4.52\% & 0 \\
MetricX & +0.28\% & +2.25\% & \textbf{+13.3\%} & -0.16\% \\
\bottomrule
\end{tabular}
}
\caption{\textbf{Translation quality with individual QE versus unified QE metrics}. Values represent the relative improvement of QE-filtered data over a randomly selected (training data) baseline for English--African language translation, averaged across all test sets.}
\label{tab:indiv_qe_metrics_gain}
\end{table}

\paragraph{Aligned QE Is Best.}

We evaluate whether QE filtering should be applied: 1) in the training direction (\textit{aligned}), 2) in the opposite of training direction (\textit{reversed}) or 3) both (\textit{mean}) directions. This issue arises in back-translation workflows, where synthetic pairs may be generated and filtered in one direction, then reversed to form training examples in the opposite direction. In such cases, the QE scoring direction becomes \textit{reversed} relative to the final training direction.
To test this, we compare otherwise identical models filtered using the \textit{aligned}, \textit{reversed}, and \textit{mean} strategies defined in §\ref{sec:quality_estimation}. Results in Table~\ref{tab:directional_bias_pooled}, averaged across translation directions and test sets, show that reversed filtering substantially degrades performance, especially on MetricX ($-12.0\%$) and SSA-COMET ($-3.1\%$), while the mean strategy is closer but still generally below the aligned reference. This indicates that QE filtering \textbf{should be aligned with the final training direction}, even when synthetic data is generated through back-translation. We thus use aligned QE filtering for the remainder of the paper.

\begin{figure*}[t]
\centering
\includegraphics[width=\linewidth]{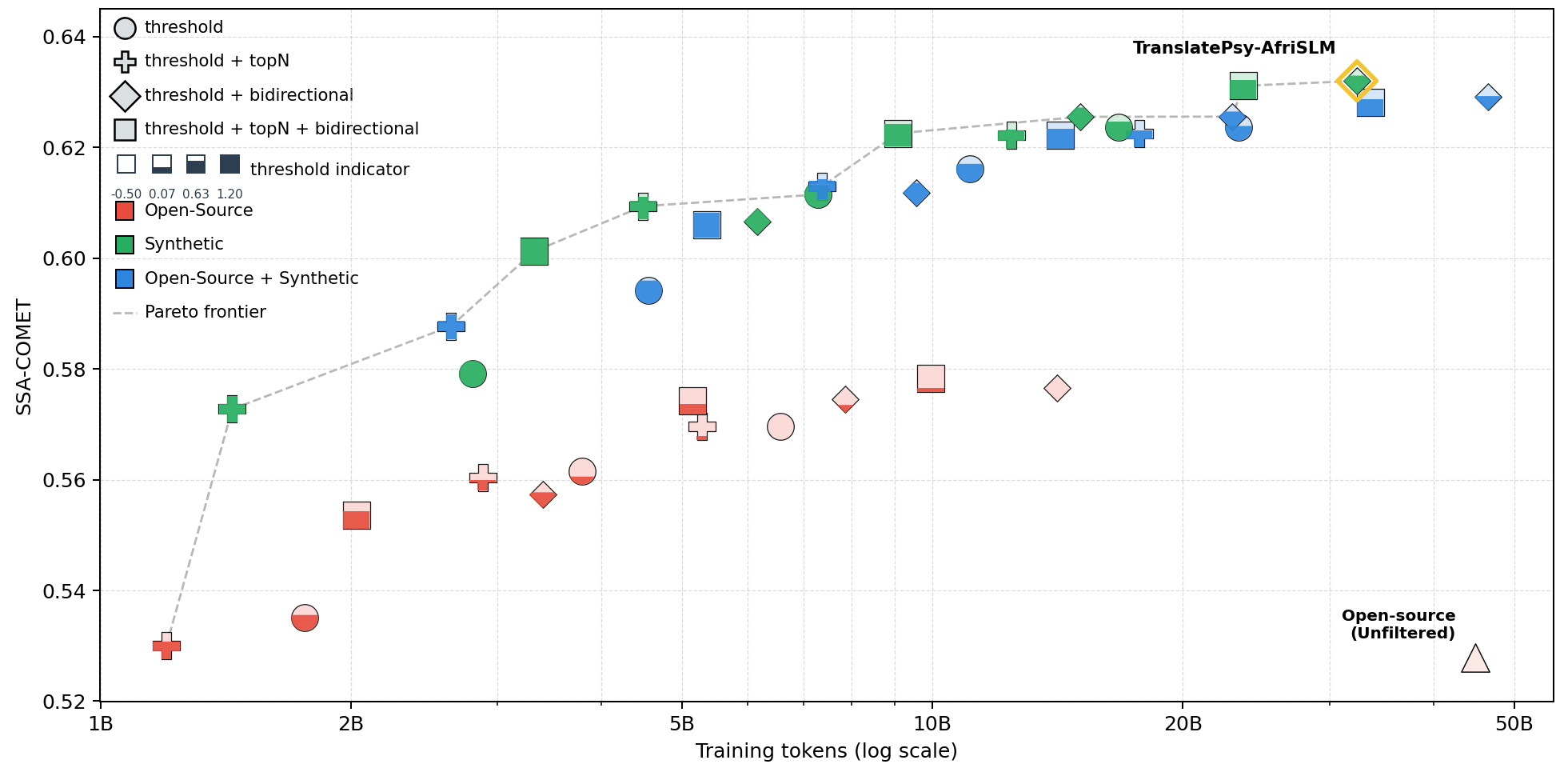}
\caption{\textbf{The Pareto frontier of performance (SSA-COMET) versus training tokens on BOUQuET.} Each shape is a fine-tuned model configuration. Colours indicate data source, shapes indicate the filtering strategy, and the shape fill indicates the threshold (the fuller the shape, the higher the threshold). Detailed plots for each metric and each dataset are shown in Figure~\ref{fig:scaling_eval_all}. The full training budgets and threshold details are reported in Table~\ref{tab:threshold_training_tokens}.}
\label{fig:scaling_pareto_plot}
\end{figure*}

\begin{table}[h]
\centering
\small
\resizebox{\columnwidth}{!}{
\begin{tabular}{lcccc}
\toprule
Direction & C22 & SSA & MX & chrF++ \\
\midrule
Aligned (Ref.) & 0 & 0 & 0 & 0 \\
Mean & -0.4\% & -0.9\% & -3.9\% & +0.7\% \\
Reversed & -1.5\% & -3.1\% & -12.0\% & +0.1\% \\
\bottomrule
\end{tabular}
}
\caption{\textbf{Training versus QE scoring directions.} Percentages show average differences to \textit{aligned} reference.}
\label{tab:directional_bias_pooled}
\end{table}
\vspace{-1em}

\subsection{How to Select Data?}
\label{sec:pareto_frontier}

With unified QE established, we study how training performance changes as a function of data quality and quantity. We sweep the z-score thresholds and combine them with the three strategies introduced in §\ref{sec:quantity} across open-source, synthetic and combined data sources. For each mix, we fine-tune Qwen3.5-2B under the same recipe. Our key observations:

\begin{enumerate}[leftmargin=*]

    % open source
    % \item \textbf{Open-source data requires aggressive quality filtering.}
    \item \textbf{Quality filtering is highly effective.}
    Figure~\ref{fig:scaling_pareto_plot} (bottom right) shows the ``Open-source (Unfiltered)'' model trained only on raw sentence pairs, totalling 44.93B tokens. Its lower relative quality can clearly be seen in Figure~\ref{fig:qe_heatmap}. In contrast, a filtered configuration (bottom left) reaches a comparable SSA-COMET score (0.530 vs. 0.528) with only 1.76B tokens---\textbf{a 96\% reduction}. This shows that raw open-source data contains a weak training signal that can be concentrated via dedicated curation, and that scaling \emph{usable} tokens matters more than simply increasing token budgets. The breakdown of our best open-source configuration is detailed in Table~\ref{tab:open_source_mix}.

    \item \textbf{Synthetic data provides better quality and a higher volume.} Because it is generated at a larger scale, using a relatively high-quality teacher model, its unfiltered sentence pairs already benefit from higher quality scores than raw open-source pairs (Figure~\ref{fig:qe_heatmap}). This allows us to apply much stricter $\mathrm{\bar{z}}$ score thresholds while still retaining sufficient training data. In competitive configurations, open-source mixes typically require relatively permissive thresholds around $\mathrm{\bar{z} \in [-0.5, 0.5]}$, whereas synthetic mixes remain viable under substantially stricter thresholds, from $\mathrm{\bar{z} \geq 0.68}$ to $\mathrm{\bar{z} \geq 1.2}$. As a result, synthetic mixes dominate open-source mixes across nearly all training-token budgets.

    % combined
    \item \textbf{Combining open-source and synthetic mixes may help at smaller scales.}
    Combined mixes can outperform synthetic-only mixes at smaller token budgets, but this advantage generally dematerialises as the budget is increased. For example, the best large combined mix uses more tokens and a lower threshold than the final synthetic mix (46.49B tokens, $\mathrm{\bar{z} \geq 0.41}$ vs. 32.37B tokens, $\mathrm{\bar{z} \geq 0.68}$), yet performs slightly worse. This suggests that once enough high-quality synthetic data is available, adding lower-quality open-source data can dilute the training signal.

    % topN
    \item \textbf{TopN capping is associated with better token efficiency.}
    Incorporating topN with threshold-only configurations as a means to limit the influence of well-resourced language pairs tends to underperform in absolute terms although it tends to deliver a higher relative performance (per token). However, the effect is weak, therefore, restricted computational budgets should generally consider the topN capping option. 

    % bidirectional 
    \item \textbf{Bidirectional expansion as a robust strategy.}
    Across open-source, synthetic, and combined mixes, configurations with bidirectional expansion always improve compared to the equivalent configuration without expansion. This suggests that expanding selected examples to both translation directions improves coverage and keeps the data more balanced. Because filtering is still aligned with the final training direction, bidirectional expansion preserves QE reliability without the degradation caused by reversed filtering.

\end{enumerate}

\noindent Based on these observations, we choose the best-performing configuration (threshold + bidirectional expansion) as our final \textbf{\modelname} synthetic mix with 32.37B tokens, reaching a score of \textbf{0.632 SSA-COMET}.

\subsection{Which Dataset(s) Should We Use?}
\label{sec:data_source_results}

In this section, we evaluate several datasets in isolation to investigate their contribution to MT.

\begin{figure}[h]
\centering
\resizebox{\columnwidth}{!}{
\includegraphics[width=\columnwidth]{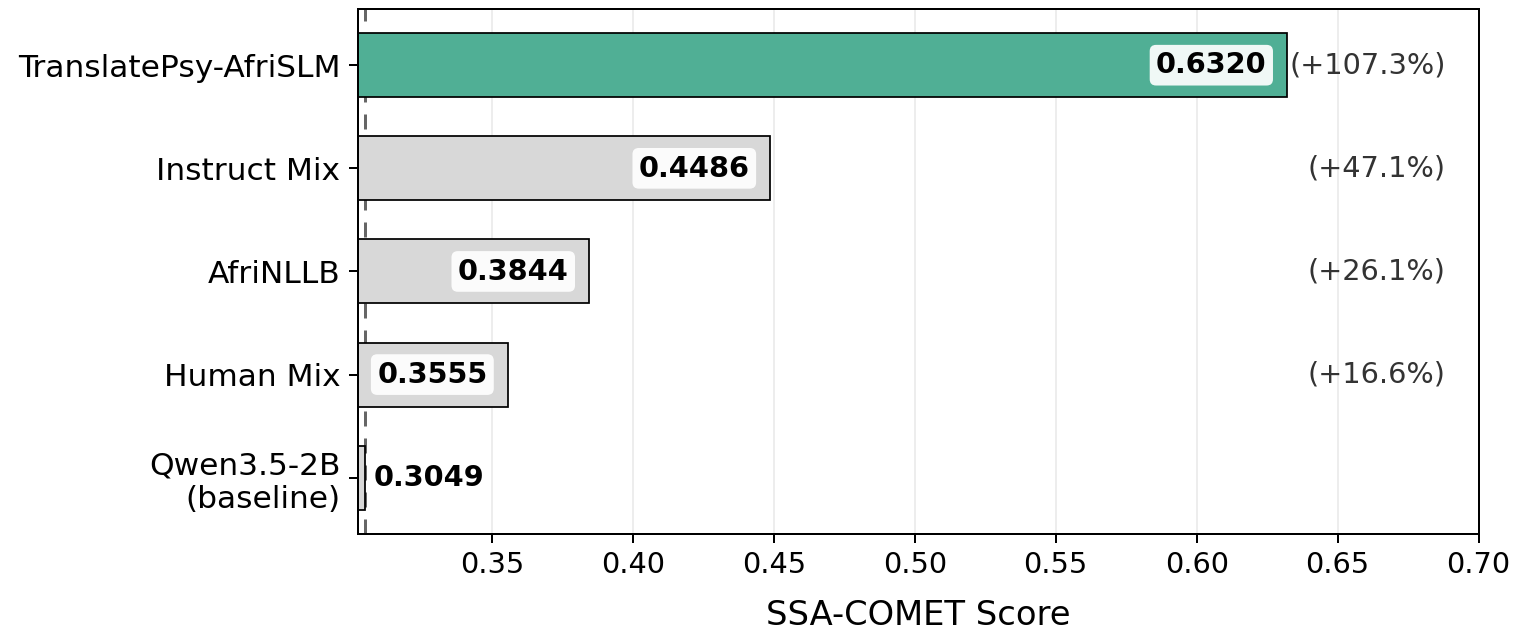}
}
\caption{\textbf{Single-source SFT ablation on BOUQuET (Qwen3.5-2B).}
Each model is fine-tuned using a single data component and evaluated with SSA-COMET. The percentages indicate relative gains over the baseline.}
\label{fig:source_ablation}
\end{figure}
\vspace{-0.5em}

\paragraph{AfriNLLB} was the largest open-source, curated dataset\footnote{Refer to Appendix \ref{sec:afrinllb_appendix} for dataset preparation.} to date, hence we positioned it as our most competitive baseline in Figure \ref{fig:source_ablation}. While it is more beneficial than the human mix, its contribution only has a moderate effect on African language MT.

\paragraph{Human Mix} is our highest-quality data, as it comes from projects with a high emphasis on human-quality translations. Figure~\ref{fig:qe_heatmap} shows that the Human Mix receives higher QE scores than AfriNLLB, which helps explain the fact that it achieves comparable performance despite being much smaller (60.2M versus 535M tokens). This suggests that human-quality data is highly valuable per token, but its limited scale makes it insufficient for effective SLM post-training on its own.

\paragraph{Instruct-Mix} The primary reason for including this dataset in the final \modelname mix is to preserve the conversational skills of our SLMs, see Figure \ref{fig:multilingual_mt_conversation} for a short demo. However, we observe that it unexpectedly benefits machine translation as well, even more so than a dedicated AfriNLLB corpus (Figure \ref{fig:source_ablation}). This is almost certainly due to the 2.3M examples (around 50\% of total) of diverse tasks in African languages. This effect is similar to the AfriqueLLM \cite{yu2026afriquellm} continued pre-training findings, which also observed an improvement in MT via instruction-following SFT.

\paragraph{\modelname} Our data\footnote{The best synthetic mix, highlighted in Figure \ref{fig:scaling_pareto_plot} (top right).} emerges as a powerful contributor to African language MT. It combines scale and quality more effectively than prior sources, retains enough data under strict filtering (Figure~\ref{fig:scaling_pareto_plot}) and achieves higher QE scores than both AfriNLLB and the Human Mix (Figure~\ref{fig:qe_heatmap}). This makes it especially effective for African language MT adaptation. The detailed statistics of this mix are given in Table~\ref{tab:synthetic_mix} in the Appendix.

\subsection{Can SLMs Beat Frontier LLMs?}
\label{sec:our_final_models}

\begin{table}[t]
\centering
\small
\resizebox{\columnwidth}{!}{
\begin{tabular}{lccc}
\toprule
\textbf{Model} & \textbf{Flores-200} & \textbf{BOUQuET} & \textbf{Smol} \\
\midrule
\multicolumn{4}{l}{\textit{\textbf{General-purpose LLMs}}} \\
\midrule
Apertus-8B & 0.3814 & 0.4024 & 0.3246 \\
% \midrule
% Qwen3.5-0.8B & 0.2967 & 0.3320 & 0.2454 \\
% Qwen3.5-2B & 0.2885 & 0.3049 & 0.2498 \\
% Qwen3.5-4B & 0.3863 & 0.3984 & 0.3301 \\
% Qwen3.5-9B & 0.4523 & 0.4664 & 0.3811 \\
Qwen3.5-27B & 0.5132 & 0.5336 & 0.4279 \\
Qwen3.5-122B-A10B & 0.5505 & 0.5716 & 0.4574 \\
\midrule
\multicolumn{4}{l}{\textit{\textbf{Dedicated translation models}}} \\
\midrule
AfriNLLB-600M & 0.5729 & 0.6038 & 0.4748 \\
% \midrule
NLLB-1.3B & 0.5878 & 0.6130 & 0.4850 \\
NLLB-3.3B & 0.5944 & 0.6178 & 0.4909 \\
% \midrule
% \midrule
Hunyuan-MT-7B & 0.4450 & 0.4451 & 0.3784 \\
% \midrule
TranslateGemma-4B & 0.3913 & 0.4069 & 0.3384 \\
% TranslateGemma-12B & 0.5037 & 0.5228 & 0.4299 \\
TranslateGemma-27B & 0.5455 & 0.5677 & 0.4608 \\
\midrule
\multicolumn{4}{l}{\textit{\textbf{African language-specialized models}}} \\
\midrule
%AfriqueLlama-8B & 0.5210 & 0.5620 & 0.4423 \\
%AfriqueGemma-12B & 0.5531 & 0.5805 & 0.4608 \\
%AfriqueQwen-14B & 0.5443 & 0.5718 & 0.4527 \\
AfriqueLlama-8B & 0.5513 & 0.5772 & 0.4555 \\
AfriqueGemma-12B & 0.5655 & 0.5892 & 0.4649 \\
AfriqueQwen-14B & 0.5569 & 0.5817 & 0.4605 \\
\midrule
\multicolumn{4}{l}{\textit{\textbf{Ours}}} \\
\midrule
\modelname-0.8B & 0.5944 & 0.6223 & 0.4973 \\
\modelname-2B & 0.6070 & 0.6322 & 0.5074 \\
\modelname-4B & \textbf{0.6143} & \textbf{0.6391} & \textbf{0.5136} \\
\bottomrule
\end{tabular}
}
\vspace{0.5em}
\caption{\textbf{\modelname compared to LLMs and specialised translators} on Flores-200, BOUQuET, Smol (SSA-COMET). Full results in tables \ref{tab:ssa_results} and \ref{tab:chrf_results}.}
\label{tab:ssa_results_short}
\end{table}

In this section, we combine \modelname with the Instruct-Mix and Asia-Europe Mix to improve real-world usability and contrast our performance with results from related methodologies.

\paragraph{Africa-IID} 
Table~\ref{tab:ssa_results_short} summarizes SSA-COMET performance across three benchmarks; full results across all four metrics are reported in Tables~\ref{tab:ssa_results} and \ref{tab:chrf_results} (Appendix~\ref{sec:additional_results_appendix}).
\textbf{(1) General-purpose LLMs} underperform, even at frontier scale. Qwen3.5-122B-A10B is surpassed by our 0.8B model on Flores-200, BOUQuET, and Smol, suggesting that model scale alone is insufficient for African language MT.
\textbf{(2) Dedicated translation models} also lag behind our SLMs. \modelname-0.8B outperforms AfriNLLB-600M and TranslateGemma-27B across all benchmarks. Notably, our 0.8B model also exceeds NLLB-3.3B on BOUQuET and Smol and matches it on Flores-200, despite using roughly a quarter of its parameters. This demonstrates that our curated post-training data can produce powerful and efficient translation SLMs.
\textbf{(3) African-centric LLMs} narrow the gap but cannot match our SLMs. AfriqueGemma-12B and AfriqueQwen-14B remain below \modelname-0.8B on the reported benchmarks, while requiring substantially more compute for training and inference.

\paragraph{Statistical significance.}
% short bersion
Paired bootstrap tests confirm that the main rankings in Table~\ref{tab:ssa_results_short} are statistically reliable (Appendix~\ref{sec:bootstrap_significance}). The results show that \modelname-0.8B significantly outperforms much larger LLM baselines on most settings, while \modelname-2B consistently surpasses dedicated NLLB models. We also observe statistically significant gains from \modelname-0.8B to 2B and 4B.

% long version
% Paired bootstrap tests confirm that the main rankings in Table~\ref{tab:ssa_results_short} are statistically reliable (Appendix~\ref{sec:bootstrap_significance}). Since the test resamples paired sentence-level scores over shared evaluation examples, it verifies that the observed gaps are robust to evaluation-set variation rather than driven by a few outlier sentences. The results show that \modelname-0.8B significantly outperforms much larger LLMs in most settings, while \modelname-2B consistently surpasses dedicated NLLB models. We also observe statistically significant gains from \modelname-0.8B to 2B \& 4B.

\begin{figure}[t]
  \centering
  \includegraphics[width=\linewidth]{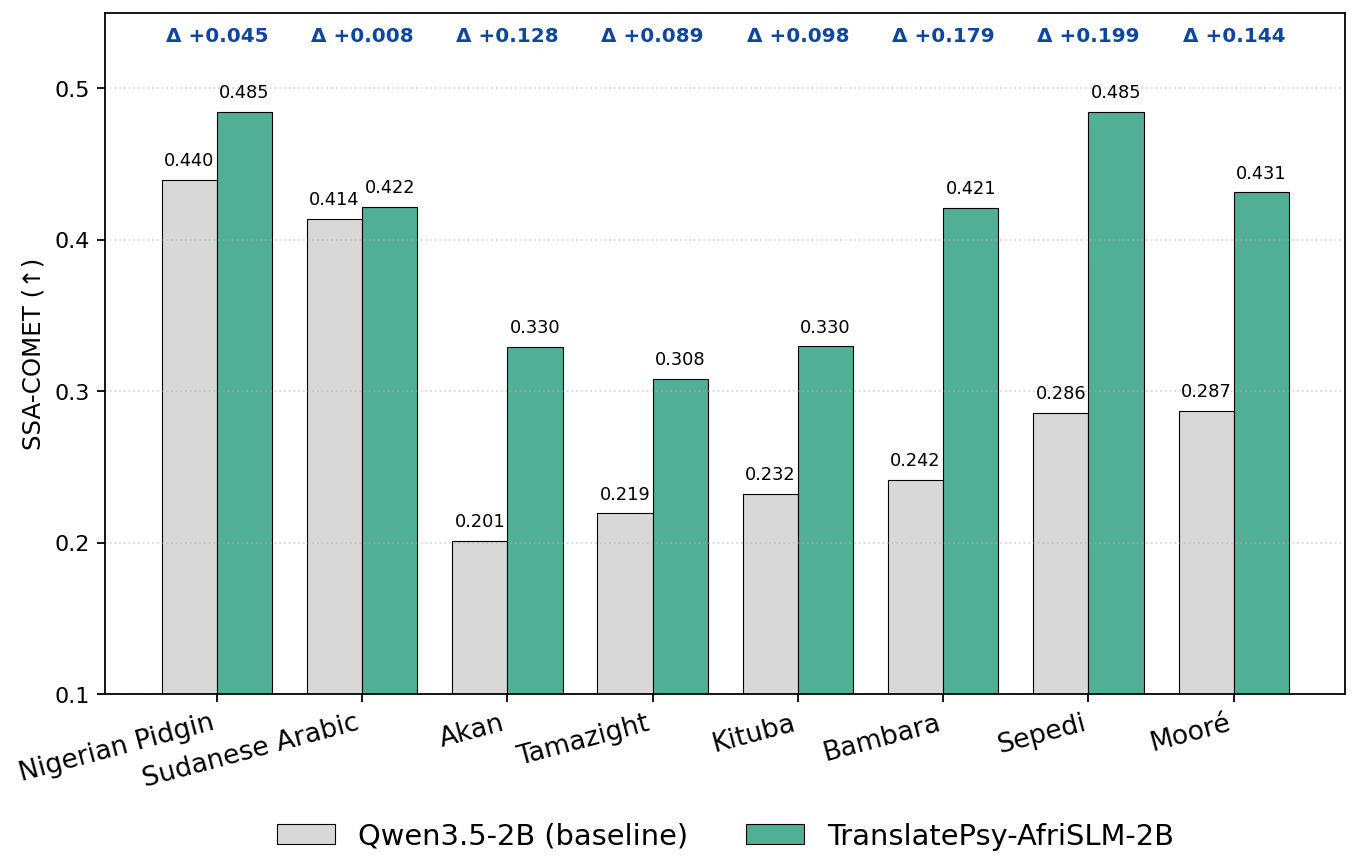}
  % \vspace{-2em}
  \caption{\textbf{Africa-OOD performance (SSA-COMET)}. Qwen3.5-2B versus \modelname-2B. Scores are averaged over translation directions and all datasets.}
  % \vspace{-0.5em}
  \label{fig:africa-ood-ssa}
\end{figure}

\paragraph{Metric circularity}
Since our fine-tuning data was filtered using quality estimators which overlap with our evaluation metrics, we acknowledge the potential for a circular evaluation bias. In order to ensure that this does not artificially distort model rankings, we confirm that held-out lexical metrics (chrF++ and spBLEU) strongly correlate with our primary evaluation metrics (Table~\ref{tab:spearman_correlation}). Across system rankings, neural and surface-level metrics \textbf{move in tight alignment} (all $p \ll 0.001$). Because these rankings hold under metrics outside of the filtering family, our performance gains are highly unlikely to be an artifact of metric-specific optimization.

\begin{table}[htbp]
\centering
\resizebox{\columnwidth}{!}{%
\begin{tabular}{l c c c c c}
\toprule
\textbf{Spearman $\rho$} & \textbf{C22} & \textbf{chrF++} & \textbf{SSA} & \textbf{MX} ($\downarrow$) & \textbf{spBLEU} \\
\midrule
\textbf{C22}             & 1   & 0.984 & 0.967 & $-0.978$ & 0.971 \\
\textbf{chrF++}          & --- & 1     & 0.958 & $-0.968$ & 0.989 \\
\textbf{SSA}       & --- & ---   & 1     & $-0.985$ & 0.952 \\
\textbf{MX} ($\downarrow$) & --- & ---   & ---   & 1     & $-0.970$ \\
\textbf{spBLEU}          & --- & ---   & ---   & ---   & 1     \\
\bottomrule
\end{tabular}%
}
\caption{Spearman correlation ($\rho$) matrix.}
\label{tab:spearman_correlation}
\end{table}

\begin{figure}[h]
\centering
\includegraphics[width=\columnwidth]{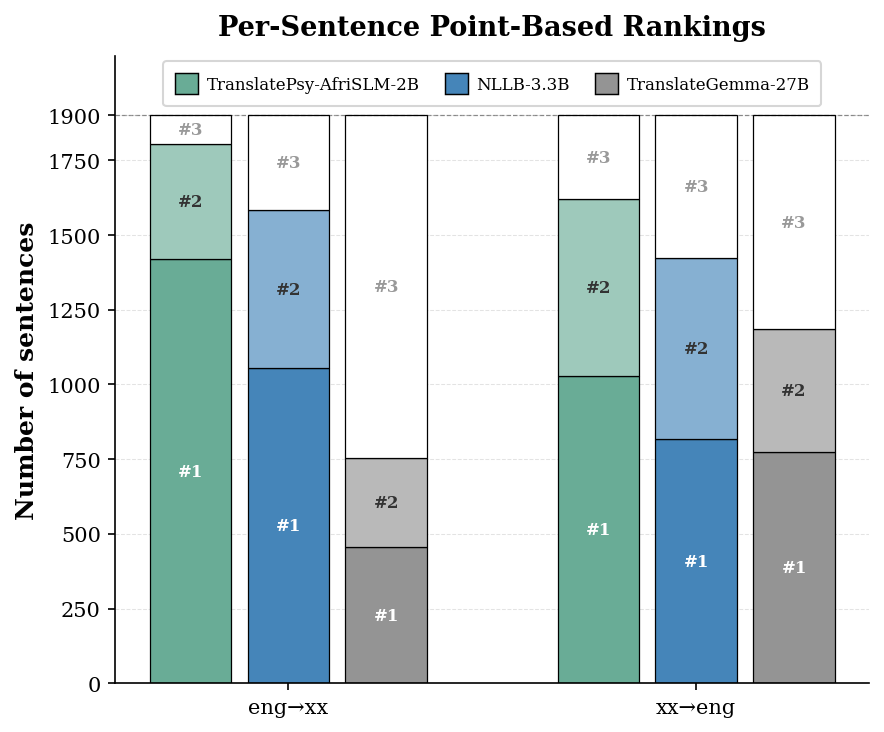}
\caption{\textbf{LLM-as-a-judge results.} Showing the number of times each model was ranked \#1 (darker shade), \#2 (lighter) or \#3 (gap up to the dotted line) for into-English and from-English language pairs.}
\label{fig:llm_as_a_judge_aggregated}
\end{figure}

\paragraph{LLM-as-a-judge}
We present further evidence against any ranking distortions by the means of evaluating \modelname-2B against NLLB-3.3B and TranslateGemma-27B using LLM-as-a-judge. We used two frontier models, GPT-5.5 and Claude Opus 4.8, to provide an independent, semantic assessment of model performance. For each language pair, we randomly selected 100 sentences from the BOUQuET test set for which the three assessed models had different outputs. We asked the judge LLMs to perform pairwise evaluations, amounting to 300 assessments for each language pair and each judge. Figure~\ref{fig:llm_as_a_judge_aggregated} plots the number of times each engine was ranked \#1, \#2 or \#3, for all  into-English and from-English language pairs (1900 sentences in each direction).  Figure~\ref{fig:llm_as_a_judge_aggregated} shows that the \textbf{rankings match automatic evaluation metric rankings} (Tables~\ref{tab:ssa_results} and \ref{tab:chrf_results}). Full details (including per language) are shown in \S\ref{sec:llm_as_a_judge}.

%\vspace{-0.5em}

\paragraph{Africa-OOD}
We use this eight language subset to evaluate if training on our 19 target languages transfers MT to unseen languages. Under SSA-COMET, \modelname-2B improves over Qwen3.5-2B on all eight held-out languages, with particularly large gains for lower-resource languages such as Sepedi, Bambara, and Akan, shown in Figure~\ref{fig:africa-ood-ssa}. Further per-language analyses show similar trends across COMET-22, MetricX, and chrF++ (Figure~\ref{fig:app-africa-ood-all-metrics}; Appendix~\ref{app:per-language}). These results suggest meaningful cross-lingual transfer beyond the training data but less uniform than IID gains.

\paragraph{Catastrophic Forgetting}

\begin{figure}[t]
\centering
\includegraphics[width=\columnwidth]{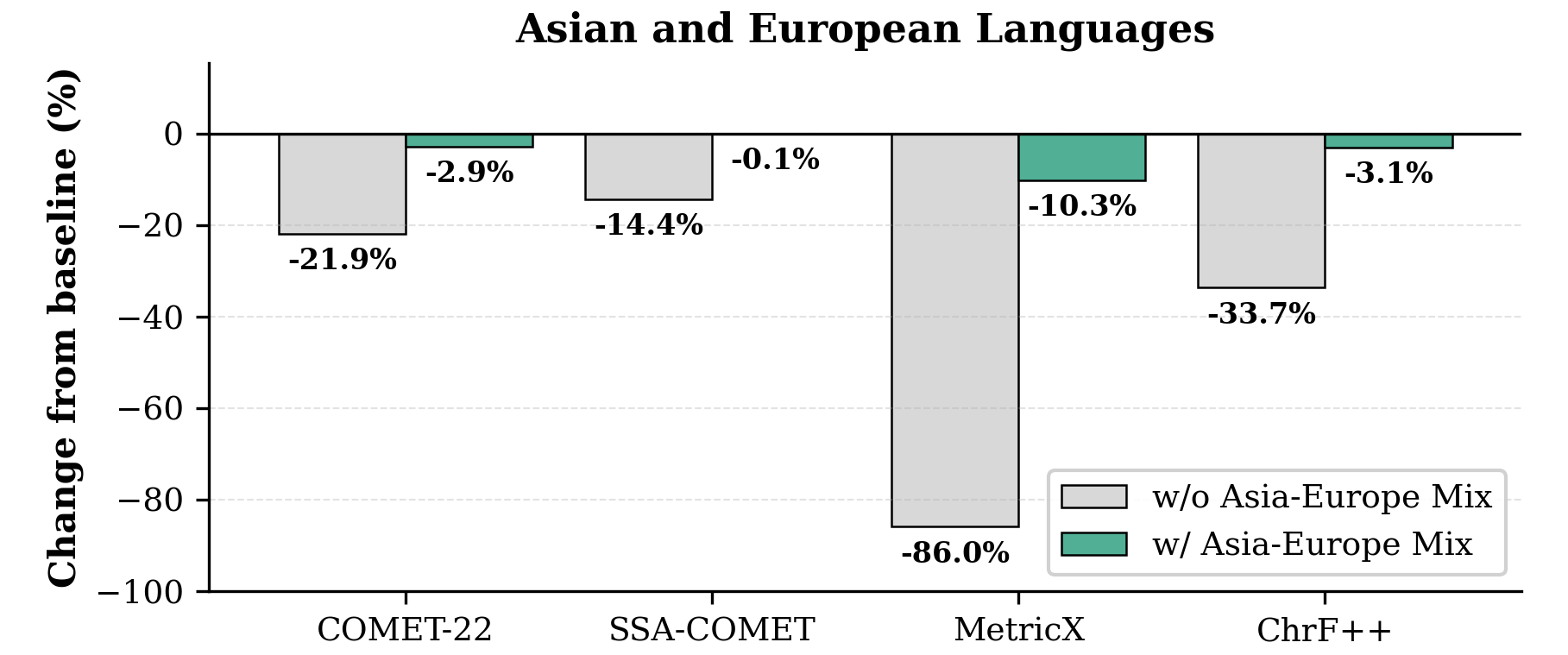}
\caption{\textbf{Catastrophic forgetting on 38 Asian and European languages.} 
The bars show the percentage change from baseline after SFT, with/without the mix.}
\label{fig:opus-forgetting}
\end{figure}

We include the Asia-Europe Mix (\S\ref{sec:additional_mixes}) to mitigate catastrophic forgetting on non-African languages. Figure~\ref{fig:opus-forgetting} shows that SFT without this data causes substantial degradation on Asian and European languages, most notably under MetricX ($-86.0\%$). Adding our Asia-Europe Mix reduces this to $-10.3\%$ and also mitigates degradation measured by COMET-22, SSA-COMET, and chrF++. These results indicate that non-African parallel data helps preserve broader multilingual translation ability without sacrificing Africa-IID performance; in fact, we observe a small improvement on Africa-IID pairs (Appendix~\ref{app:catastrophic-forgetting}).

\vspace{-0.5em}

\section{Conclusions}

We have introduced \textbf{\modelname}, a comprehensive investigation of training data curation aimed at improving machine translation for 19 Sub-Saharan languages. 
Our results lead to two main conclusions: 1) Combining multiple QE metrics with robust $\bar{z}$ score normalization, and scoring examples in the same direction as training, yields more consistent filtering than relying on any single metric. This allows us to reduce the number of training tokens by up to 96\% while maintaining comparable performance, 2) Across the 1B--50B token range, filtered synthetic data dominates the quality-efficiency Pareto frontier, while open-source data appears to saturate early due to quality limitations. Given the current shortage of high-quality open-source parallel data for African language MT, filtered synthetic generation seems to be the most practical path forward. Guided by these findings, we post-trained \modelname SLMs on our highest-quality data, outperforming much larger models, including TranslateGemma-27B and Qwen3.5-122B-A10B, with as few as 0.8B parameters. We open-source\footnote{\url{https://huggingface.co/collections/qvac/translatepsy-afrislm}} our data to support future work on making African language MT a default capability in multilingual models.

% \begin{ack}
% All acknowledgments go at the end of the paper before the list of references. Moreover, you are required to declare funding (financial activities supporting the submitted work) and competing interests (related financial activities outside the submitted work). Do {\bf not} include this section in the anonymized submission, only in the final paper. You can use the \texttt{ack} environment provided in the style file to automatically hide this section in the anonymized submission.
% \end{ack}

\section*{Limitations}

\paragraph{The Need for Human Evaluation.} One notable finding that warrants future investigation is the challenge of determining absolute translation quality for African languages. Using the \textbf{reference-free quality estimators} from prior work, it appears that our \modelname data is comparable to human translated pairs in quality but it comes in quantities that are orders of magnitude larger. Still, the performance, as indicated by the \textbf{reference-based evaluation metrics} (produced by same model families as QE) suggests that we have not yet reached the translation quality of European and Asian languages. Therefore, conducting purely quantitative data scaling efforts using existing tools may run into \textbf{unknown absolute performance limitations}. The reason(s) behind this can only be resolved by analysing each step in the pipeline [data collection and training of QE models, correlation with human judgement, data collection and training of evaluation models, their correlation with ground truth], \textbf{leveraging expert human annotators} to identify any errors that have the capacity to compound over multiple steps during a large-scale data curation. 
A specific target for human evaluation is investigating the discrepancy between covered and uncovered languages within current QE estimators. Although we filter training data across all language pairs evaluated in this work, these estimators were trained only on a subset of those. A study on the impact of filtering languages absent from QE training would be an insightful direction for future work.

\section*{Ethics Statement}
\paragraph{Data licensing} All primary sources are open-source and publicly available under permissive attribution-based licenses — ODC-By v1.0 (Fine Translations, MADLAD-400) and CC BY 4.0 (MaLA Corpus), alongside research-permissive aggregates from OPUS and WMT22. The synthetic \modelname mix will be released under the CC-BY-NC 4.0 license, for safe, lawful, and reproducible low-resource NLP research.
\paragraph{Dialectal representativeness} The exact dialect tracking is unavailable for our web-curated corpora; models may over-reflect standardized written forms and under-represent regional dialects or oral traditions. We advise auditing trained models for localized sensitivity before deployment.
\paragraph{Synthetic content} In our "open-source" mix, the Fine Translations corpus is entirely synthetic. Because large internet repositories involve data recycling, mixing, duplication, and some LLM-based corrections and paraphrasing, an unknown portion of the "open-source" subset should be treated as at least partially synthetic. The open-source mix will not be released due to licensing/provenance issues.

\bibliography{references}
\bibliographystyle{acl_natbib}

\appendix
\newpage
\appendix

\section{Appendix}
\label{sec:appendix}

\subsection{Synthetic and Open-Source Mixes}
\label{appendix:data_processing}

The corpora composing these mixes are described in §\ref{sec:data_sources}. Here, we give more details of the processing steps and data statistics. The preprocessing steps (up to decontamination) were implemented with NeMo Curator~\cite{nemo_curator}. In Tables \ref{tab:open_source_mix} and \ref{tab:synthetic_mix}, "Clean" means before deduplication, "Preproc" after decontamination, "Filtered" after QE filtering, and "Bidir." after bidirectional expansion.

\paragraph{Processing steps:}
\begin{enumerate}[nosep]
    \item \textbf{Document Splitting:}\footnote{\label{pipeline_splitting}Document and sentence splitting was performed only on monolingual source texts to be translated for synthetic parallel data generation, because the teacher is not accurate/designed for paragraph-long input texts.} has divided documents into paragraphs (splitting at newlines).
    \item \textbf{Cleaning:} removed Unicode artifacts, extra newlines, markup, and URLs.
    \item \textbf{Language Identification:} via \textit{AfroLID} \cite{adebara-etal-2022-afrolid} for African languages and \textit{FastText}~\cite{joulin2016bag} for English.
    \item \textbf{Sentence Splitting:}\footnotemark[\getrefnumber{pipeline_splitting}] was performed using pySBD~\cite{sadvilkar-neumann-2020-pysbd}.
    \item \textbf{Filtering:} removed non-alphanumeric content and boilerplate text.
    \item \textbf{Deduplication:} with exact \& fuzzy matching.
    \item \textbf{Eval-set Decontamination:} removed sentences similar to our test sets (§\ref{sec:decontamination}).
    \item \textbf{Synthetic Data Generation} The monolingual data (§\ref{sec:data_sources}) was translated with the NLLB-3.3B model,\footnote{\scriptsize\url{https://huggingface.co/facebook/nllb-200-3.3B}} decoded with the c-translate2 decoder\footnote{\scriptsize\url{https://github.com/OpenNMT/CTranslate2}} with "int8\_float16" quantisation and beam size 3. We discarded the 54B MoE model because the MoE architecture is not supported by fast decoders.

    \item \textbf{Quality Estimation (QE):} For open-source data, the input to the QE stage is the parallel pre-processed text. For the synthetic data, it is the monolingual pre-processed source text together with its translation. For each pair, MetricX, AfriCOMET, SSA-COMET scores and the z-score were calculated in the source-target and target-source directions. Aligned filtering was applied (§\ref{sec:quality_estimation}), and pairs with a z-score below the threshold were discarded. 
\end{enumerate}

\begin{table}[h]
\centering
\small
\resizebox{\columnwidth}{!}{%
\begin{tabular}{@{}l*{7}{r}@{}}
\toprule
 &  &  &  & \multicolumn{4}{c}{QE Filtered} \\
\cmidrule(lr){5-8}
Lang. & Raw & Clean & Preproc & en$\to$xx & xx$\to$en & xx$\to$yy & Bidir. \\
\midrule
afr & 66,089 & 48,932 & 43,060 & 1,500 & 1,000 & 2,248 & 7,425 \\
amh & 41,109 & 36,935 & 30,542 & 1,020 & 776 & 2,841 & 6,518 \\
hau & 33,515 & 24,380 & 22,537 & 1,230 & 42 & 1,453 & 4,162 \\
ibo & 17,550 & 15,378 & 14,385 & 1,500 & 17 & 309 & 3,458 \\
kin & 21,855 & 16,425 & 15,484 & 1,500 & 0 & 280 & 2,989 \\
lin & 9,992 & 6,169 & 5,727 & 1,500 & 1 & 0 & 2,217 \\
lug & 14,257 & 8,175 & 7,528 & 1,302 & 4 & 80 & 1,922 \\
mlg & 11,106 & 8,336 & 7,829 & 1,394 & 10 & 19 & 2,477 \\
nya & 14,655 & 10,715 & 9,804 & 586 & 14 & 537 & 1,952 \\
orm & 11,949 & 5,703 & 5,517 & 892 & 8 & 7 & 1,175 \\
sna & 21,477 & 12,676 & 11,744 & 1,391 & 11 & 510 & 3,488 \\
som & 26,157 & 22,432 & 20,068 & 1,388 & 39 & 292 & 3,425 \\
sot & 4,517 & 3,676 & 3,491 & 23 & 1 & 0 & 55 \\
swa & 37,664 & 30,784 & 28,223 & 1,500 & 41 & 697 & 5,947 \\
tsn & 14,595 & 9,325 & 8,520 & 1,609 & 25 & 63 & 3,085 \\
wol & 4,342 & 3,506 & 3,280 & 868 & 1 & 14 & 1,218 \\
xho & 22,971 & 14,435 & 13,534 & 1,500 & 13 & 200 & 3,595 \\
yor & 13,909 & 9,052 & 8,091 & 411 & 67 & 69 & 1,999 \\
zul & 24,476 & 13,709 & 12,950 & 1,132 & 15 & 3 & 3,196 \\
others & 15,004 & 14,828 & 14,292 & 1,500 & 1,000 & 0 & 5,000 \\
\midrule
\textbf{Total} & 427,187 & 315,571 & 286,606 & 23,747 & 3,085 & 9,623 & 65,304 \\
\bottomrule
\end{tabular}
}
\caption{\textbf{Open-Source mix} (line counts in thousands).}
\label{tab:open_source_mix}
\end{table}

\begin{table}[h]
\centering
\resizebox{\columnwidth}{!}{%
\begin{tabular}{@{}l*{6}{r}@{}}
\toprule
 & \multicolumn{3}{c}{Monolingual} & \multicolumn{3}{c}{Parallel, QE Filtered} \\
\cmidrule(lr){2-4} \cmidrule(lr){5-7}
Lang. & Raw & Clean & Preproc & eng$\to$xx & xx$\to$eng & Bidir. \\
\midrule
eng & 145,726 & 145,688 & 136,482 & --- & --- & --- \\
afr & 24,062 & 23,384 & 11,139 & 1,009 & 1,875 & 6,974 \\
amh & 3,704 & 3,576 & 3,379 & 4,897 & 569 & 10,786 \\
hau & 5,002 & 4,729 & 4,420 & 6,697 & 1,306 & 17,261 \\
ibo & 2,450 & 2,351 & 2,222 & 5,964 & 639 & 12,659 \\
kin & 5,211 & 4,699 & 4,219 & 2,536 & 847 & 8,102 \\
lin & 153 & 145 & 135 & 5,056 & 36 & 7,482 \\
lug & 378 & 358 & 332 & 5,586 & 48 & 10,089 \\
mlg & 3,521 & 2,690 & 2,464 & 1,010 & 470 & 4,168 \\
nya & 2,292 & 1,846 & 1,768 & 3,575 & 439 & 8,060 \\
orm & 75 & 70 & 68 & 4,817 & 10 & 7,580 \\
sna & 426 & 344 & 335 & 2,490 & 92 & 5,829 \\
som & 6,981 & 6,616 & 5,967 & 2,744 & 851 & 7,381 \\
sot & 1,980 & 1,851 & 1,754 & 4,576 & 503 & 8,495 \\
swa & 18,080 & 17,559 & 15,834 & 3,598 & 2,776 & 17,670 \\
tsn & 22 & 21 & 20 & 3,506 & 7 & 6,033 \\
wol & 45 & 43 & 39 & 13,256 & 7 & 23,615 \\
xho & 2,219 & 2,040 & 1,915 & 4,414 & 465 & 7,962 \\
yor & 2,345 & 2,109 & 2,007 & 13,464 & 1,128 & 27,140 \\
zul & 2,118 & 1,737 & 1,662 & 9,115 & 519 & 18,367 \\
\midrule
\textbf{Total} & 226,790 & 221,856 & 196,162 & 98,312 & 12,589 & 215,653 \\
\bottomrule
\end{tabular}
}
\caption{\textbf{Synthetic-Mix:} i.e. the \modelname training data by language (line counts in thousands).}
\label{tab:synthetic_mix}
\end{table}

\begin{table}[t]
\centering
\small
\begin{tabular}{lcccc}
\toprule
Quality estimator & C22 & SSA & MX & chrF++ \\
\midrule
\multicolumn{5}{c}{eng-xx (\textit{English $\rightarrow$ African language(s))}} \\
\midrule
Random & 0.727 & 0.590 & 5.22 & 45.1 \\
z-score & \textbf{0.737} & 0.617 & 4.52 & 45.7 \\
SSA-COMET & 0.736 & \textbf{0.620} & 4.73 & \textbf{45.9} \\
AfriCOMET & 0.735 & 0.609 & 4.87 & 45.3 \\
MetricX & 0.730 & 0.605 & \textbf{4.29} & 45.2 \\
\midrule
\multicolumn{5}{c}{xx-eng (\textit{African language(s) $\rightarrow$ English})} \\
\midrule
Random & 0.741 & 0.566 & 6.20 & \textbf{52.9} \\
z-score & 0.754 & 0.581 & 5.30 & 52.0 \\
SSA-COMET & 0.746 & \textbf{0.581} & 5.79 & 51.9 \\
AfriCOMET & \textbf{0.756} & 0.579 & 5.40 & 52.3 \\
MetricX & 0.749 & 0.573 & \textbf{5.27} & 51.7 \\
\bottomrule
\end{tabular}
\caption{\textbf{Translation quality with individual versus unified QE metrics} over Flores-200 test set.}
\label{tab:indiv_qe_metrics_flores}
\end{table}

\begin{table}[t]
\centering
\small
\begin{tabular}{lcccc}
\toprule
Quality estimator & C22 & SSA & MX & chrF++ \\
\midrule
\multicolumn{5}{c}{eng-xx (\textit{English $\rightarrow$ African language(s))}} \\
\midrule
Random & 0.646 & 0.476 & 9.59 & 27.8 \\
z-score & \textbf{0.655} & 0.503 & 8.88 & \textbf{28.0} \\
SSA-COMET & 0.652 & \textbf{0.505} & 9.22 & \textbf{28.0} \\
AfriCOMET & 0.653 & 0.495 & 9.26 & 27.8 \\
MetricX & 0.650 & 0.491 & \textbf{8.58} & 27.9 \\
\midrule
\multicolumn{5}{c}{xx-eng (\textit{African language(s) $\rightarrow$ English})} \\
\midrule
Random & 0.563 & 0.482 & 10.93 & \textbf{30.1} \\
z-score & 0.570 & \textbf{0.500} & 10.25 & 29.1 \\
SSA-COMET & 0.565 & 0.498 & 10.65 & 29.2 \\
AfriCOMET & \textbf{0.570} & 0.497 & 10.40 & 29.3 \\
MetricX & 0.568 & 0.491 & \textbf{10.14} & 29.2 \\
\bottomrule
\end{tabular}
\caption{\textbf{Translation quality with individual versus unified QE metrics} over the Smol test set.}
\label{tab:indiv_qe_metrics_smol}
\end{table}

\newpage

\subsection{Human Mix}
\label{human}
We compiled 352,582 human-translated, high-quality parallel sentences focused on African languages from AfriDOC-MT \cite{alabi2025afridoc} and SMOL\footnote{Excluding our evaluation set (\textbf{smolsent}).} \cite{caswell2025smol}. Table~\ref{tab:human-mix-smoldoc} summarizes the high-level statistics. The SmolDoc subset of SMOL\footnote{ \url{https://huggingface.co/datasets/google/smol}} was flattened to sentence-level pairs.

\paragraph{Processing Steps:}
\begin{enumerate}[nosep]
    \item \textbf{Bidirectional expansion:} Both directions (eng$\rightarrow$xx and xx$\rightarrow$eng) were used for training.
    \item \textbf{Global exact deduplication:} Identical pairs were removed.
    \item \textbf{Eval-set decontamination:} Removed sentences similar to our evaluation data.
    \item \textbf{Global approximate deduplication:} MinHash applied to remove similar examples.
\end{enumerate}

\begin{table}[H]
\centering
\small
\resizebox{\columnwidth}{!}{
\begin{tabular}{lrrr}
\toprule
Filter Type & Kept & Removed & Retained \% \\
\midrule
Raw sentence pairs & 180,540 & -- & 100.00\% \\
(1) After bidirectional expansion & 361,080 & -- & 200.00\% \\
(2) After global exact deduplication & 357,584 & 3,496 & 99.03\% \\
(3) After eval-set decontamination & 357,402 & 182 & 99.95\% \\
(4) After global approximate deduplication & 352,582 & 4,820 & 98.65\% \\
\bottomrule
\end{tabular}
}
\caption{Filtering summary for the human translations.}
\label{tab:human-mix-smoldoc}
\end{table}

\subsection{Instruct Mix}
\label{sec:instruction_mix_appendix}

\paragraph{Afri-Instruct Mix}
This dataset is a heterogeneous mixture of open-ended instruction data aggregated from 11 publicly available HuggingFace datasets  \cite{singh2024aya,alpaca,devine2026kakugo,owusu2025africode_collection,muhammad2023semeval,Adelani2023MasakhaNEWS,ogundepo2023afriqa,adelani2025irokobench,ojo2025afrobench,hasan-etal-2021-xl,ding2023enhancing}. Table~\ref{tab:afri-instruct-sources} summarizes each data source. The final mixture contains 2,306,800 examples designed to maintain/extend essential instruction-following abilities to African language MT, e.g. open-ended instruction and chat data, which contributes 1,553,944 examples (67.36\%), followed by code-assistance data with 521,389 examples (22.60\%). Smaller portions come from classification and other structured prediction tasks with 123,448 examples (5.35\%), summarization and headline generation with 74,581 examples (3.23\%), and question answering plus cross-lingual QA with 33,438 examples (1.45\%).

\vspace{2mm}
\paragraph{Processing Steps:}

\begin{enumerate}[nosep]
    \item \textbf{Format standardization:} Each dataset was converted to a multi-turn chat format with system, user, and assistant roles. Appropriate system prompts were added (e.g., ``You are a sentiment classifier'' for AfriSenti, ``You are a summarization assistant'' for XLSum), etc.
    \item \textbf{Quality filtering:} Samples with empty user or assistant messages were removed.
    \item \textbf{Task expansion:} AfriQA was expanded into three task types: machine translation pairs, monolingual QA, and cross-lingual QA, tripling its effective sample count.
    \item \textbf{Sampling:} For Afri-Code datasets, a maximum of 30,000 samples per language were randomly sampled to reduce repetition.
\end{enumerate}

\begin{table}[htbp]
\centering
\small
\resizebox{\columnwidth}{!}{
\begin{tabular}{llcc}
\toprule
\textbf{Dataset} & \textbf{Source} & \textbf{Count} & \textbf{Languages} \\
\midrule
African-UltraChat & \texttt{masakhane/african-ultrachat} & 54,994 & 11 \\
African-Alpaca & \texttt{masakhane/african-translated-alpaca} & 832,029 & 16 \\
Kakugo & \texttt{ptrdvn/kakugo-\{lang\}} & 464,569 & 12 \\
Aya Dataset & \texttt{CohereLabs/aya\_dataset} & 202,352 & 65 \\
Afri-Code & \texttt{michsethowusu/Code-170k-\{lang\}} & 521,389 & 18 \\
AfriSenti & \texttt{shmuhammad/AfriSenti-twitter-sentiment} & 83,688 & 8 \\
MasakhaNEWS & \texttt{masakhane/masakhanews} & 21,296 & 12 \\
XLSum & \texttt{csebuetnlp/xlsum} & 53,285 & 7 \\
AfriADR & \texttt{masakhane/AfriADR} & 26,110 & 3 \\
AfriXNLI & \texttt{masakhane/afrixnli} & 13,650 & 13 \\
AfriQA & \texttt{masakhane/afriqa} & 33,438 & 7 \\
\midrule
\textbf{Total} & - & \textbf{2,306,800} & - \\
\bottomrule
\end{tabular}
}
\caption{Data sources for the Afri-Instruct data mix.}
\label{tab:afri-instruct-sources}
\end{table}

\paragraph{General-Instruct Mix}
Additional instruction-following training examples (mostly English-centric) come from two public HuggingFace datasets: \textbf{smoltalk2}, the post-training data of SmolLM3 \cite{bakouch2025smollm3} and \textbf{Dolci-Instruct}, the OLMO-3 supervised fine-tuning data \cite{olmo2025olmo3}. Table~\ref{tab:instruct-mix-sources} summarizes each data source. Exact deduplication removed 85,982 examples. Due to the exclusion of reasoning data in \modelname training, it does not retain the 'thinking' capability of its base model.

\paragraph{Processing Steps:}
\begin{enumerate}[nosep]
    \item \textbf{Format standardization:} Datasets were converted to a unified chat format. A system prompt (``You are a helpful assistant.'') was added to conversations from Dolci-Instruct.
    \item \textbf{Exact deduplication:} Identical examples were removed.
\end{enumerate}

\begin{table}[htbp]
\centering
\small
\resizebox{\columnwidth}{!}{
\begin{tabular}{lll}
\toprule
\textbf{Dataset} & \textbf{Source} & \textbf{Splits} \\
\midrule
SmolTalk2 & \texttt{HuggingFaceTB/smoltalk2} (SFT) & \makecell[l]{
    multilingual\_8languages\_lang\_5\_no\_think \\
    smollm3\_systemchats\_30k\_no\_think \\
    smollm3\_everyday\_conversations\_no\_think \\
    smollm3\_explore\_instruct\_rewriting\_no\_think \\
    smollm3\_smol\_rewrite\_no\_think \\
    smollm3\_smol\_summarize\_no\_think
} \\
\midrule
Dolci-Instruct & \texttt{allenai/Dolci-Instruct-SFT-No-Tools} & train \\
\midrule
\textbf{Total} & - & \textbf{2,308,569} \\
\bottomrule
\end{tabular}
}
\caption{\centering Data sources for the General Instruct mix.}
\label{tab:instruct-mix-sources}
\end{table}

\subsection{Asia-Europe Mix}
\label{sec:europe_asia_appendix}

In order to mitigate catastrophic forgetting, we included a broad selection of 38 medium-high resource languages\footnote{\scriptsize \url{https://huggingface.co/datasets/Helsinki-NLP/opus-100}} from OPUS-100 \cite{zhang-etal-2020-improving,tiedemann-2012-parallel}. The final dataset comprises 24,114,303 parallel sentences. We only use en $\rightarrow$ xx pairs as the xx $\rightarrow$ en direction is robust to large-scale African language post-training. Including this data has negligible impact on African language translation quality, while it plays a crucial role in preserving Asian and European language performance (Appendix~\ref{app:catastrophic-forgetting}). Table \ref{tab:europe_asia} provides an aggregate summary while Table~\ref{tab:opus-mix-sources} in summarizes the per-language statistics.

\paragraph{Processing Steps:}
\begin{enumerate}[nosep]
    \item \textbf{Quality filtering:} Samples were filtered to only those with a mean bidirectional COMET\footnote{\scriptsize \url{https://huggingface.co/Unbabel/wmt22-cometkiwi-da}} \cite{rei-etal-2022-cometkiwi} score $\geq 0.6$.
    \item \textbf{Approximate deduplication:} MinHash was applied using 128 permutations and a Jaccard similarity of 0.8 on character 4-grams.
    \item \textbf{Eval-set decontamination:} Sentences similar to our evaluation sets were removed.
    \item \textbf{Global approximate deduplication:} BPE-unigram MinHash deduplication (256 permutations, Jaccard threshold 0.8) was applied globally to remove similar examples.
\end{enumerate}

\begin{table}[h]
\small
\centering
\resizebox{\columnwidth}{!}{
\begin{tabular}{lrrr}
\toprule
\textbf{Filter Type} & \textbf{Kept} & \textbf{Removed} & \textbf{Retained} \% \\
\midrule
Raw sentence pairs & 36,347,268 & -- & 100.00\% \\
After COMET mean $\ge 0.6$ & 30,112,876 & 6,234,392 & 82.85\% \\
After pair-level (local) MinHash deduplication & 25,439,466 & 4,673,410 & 84.48\% \\
After eval-set decontamination & 25,429,129 & 10,337 & 99.96\% \\
After MinHash deduplication & 24,114,303 & 1,314,826 & 94.83\% \\
\bottomrule
\end{tabular}
}
\caption{Filtering summary for Europe/Asia mix.}
\label{tab:europe_asia}
\end{table}

\begin{table}[h]
\centering
\small
\resizebox{\columnwidth}{!}{
\begin{tabular}{lrrr|lrrr}
\toprule
\textbf{Pair} & \textbf{RAW} & \textbf{COMET} & \textbf{DEDUP} & \textbf{Pair} & \textbf{RAW} & \textbf{COMET} & \textbf{DEDUP} \\
\midrule
en-zh & 1,000k & 837k & 767k & en-sv & 1,000k & 828k & 725k \\
en-ja & 1,000k & 788k & 669k & en-hr & 1,000k & 829k & 738k \\
en-it & 1,000k & 843k & 752k & en-is & 1,000k & 650k & 485k \\
en-ru & 1,000k & 817k & 751k & en-lt & 1,000k & 879k & 650k \\
en-es & 1,000k & 879k & 791k & en-ms & 1,000k & 815k & 634k \\
en-tr & 1,000k & 880k & 778k & en-id & 1,000k & 860k & 720k \\
en-fr & 1,000k & 839k & 789k & en-si & 979k & 829k & 475k \\
en-pl & 1,000k & 787k & 695k & en-hi & 534k & 474k & 298k \\
en-ar & 1,000k & 809k & 751k & en-bn & 1,000k & 896k & 604k \\
en-uk & 1,000k & 803k & 578k & en-ur & 754k & 727k & 603k \\
en-pt & 1,000k & 848k & 748k & en-fa & 1,000k & 813k & 730k \\
en-sk & 1,000k & 860k & 716k & en-kk & 80k & 57k & 41k \\
en-de & 1,000k & 773k & 720k & en-ro & 1,000k & 831k & 725k \\
en-hu & 1,000k & 828k & 732k & en-bg & 1,000k & 814k & 718k \\
en-el & 1,000k & 822k & 735k & en-cs & 1,000k & 814k & 720k \\
en-ko & 1,000k & 726k & 617k & en-da & 1,000k & 820k & 712k \\
en-vi & 1,000k & 863k & 729k & en-lv & 1,000k & 879k & 652k \\
en-th & 1,000k & 841k & 721k & en-nl & 1,000k & 830k & 747k \\
en-fi & 1,000k & 810k & 732k & en-et & 1,000k & 814k & 691k \\
\bottomrule
\end{tabular}
}
\caption{Language statistics for Asia-Europe data.}
\label{tab:opus-mix-sources}
\end{table}

\subsection{AfriNLLB}
\label{sec:afrinllb_appendix}

We apply only test data decontamination and minimal approximate deduplication, see Table \ref{tab:afri-nllb}.

\paragraph{Processing Steps:}
\begin{enumerate}[nosep]
    \item \textbf{Approximate deduplication:} MinHash was applied using 128 permutations and a Jaccard similarity of 0.8 on character 4-grams.
    \item \textbf{Eval-set decontamination:} Sentences similar to our evaluation sets were removed.
\end{enumerate}

\begin{table}[h]
\small
\centering
\resizebox{\columnwidth}{!}{
\begin{tabular}{lrrr}
\toprule
\textbf{Filter Type} & \textbf{Kept} & \textbf{Removed} & \textbf{Retained} \% \\
\midrule
Original AfriNLLB pairs & 3,218,822 & -- & 100.00\% \\
After eval-set decontamination & 3,206,918 & 11,904 & 99.63\% \\
After approximate deduplication & 3,129,175 & 77,743 & 97.58\% \\
\bottomrule
\end{tabular}
}
\caption{AfriNLLB preprocessing steps.}
\label{tab:afri-nllb}
\end{table}

\subsection{Data Decontamination}
\label{sec:decontamination}
As an essential part of our preprocessing steps, we decontaminate our training data against all evaluation sentences, languages and dataset splits, totaling \textbf{over 850K sentences}. We used a BPE-unigram (Qwen3.5) MinHash (LSH) deduplication with a Jaccard similarity threshold of 0.9 to filter training data on an individual sentence level rather than a pair level for the most granular detection possible.

% \subsection{Open-Source Mix}
% \label{sec:filtered_mix_appendix}

% \input{tables/table_africa_open_source}

\section{Additional Experimental Setup}

\begin{figure}[htb]
\centering
\vspace{-3.5mm}
\begin{tcolorbox}[
    colframe=black,
    colback=gray!5!white,
    title=\textbf{MT Prompt Template},
    fonttitle=\bfseries
]
\footnotesize

\textbf{System Prompt:} \\
You are a professional \texttt{SOURCE\_LANG} to \texttt{TARGET\_LANG} translator. 
Your goal is to accurately convey the meaning and nuances of the original 
\texttt{SOURCE\_LANG} text while adhering to \texttt{TARGET\_LANG} grammar, 
vocabulary, and cultural sensitivities. Produce only the \texttt{TARGET\_LANG} 
translation, without any additional explanations or commentary.

\vspace{2mm}
\noindent \textbf{User:} \\
Please translate the following \texttt{SOURCE\_LANG} text into 
\texttt{TARGET\_LANG}: \texttt{SOURCE\_TEXT}. Translation:

\vspace{2mm}
\noindent \textbf{Assistant:} \\
\texttt{TARGET\_TEXT}

\end{tcolorbox}
\vspace{-3.5mm}
\caption{Training and evaluation prompt template.}
\label{fig:translation_prompt}
\end{figure}

\subsection{Translation Prompt}

Training and evaluation examples were structured as shown in Figure \ref{fig:translation_prompt} where \texttt{SOURCE\_LANG} and \texttt{TARGET\_LANG} denote the source and target language \textbf{names} while \texttt{SOURCE\_TEXT} and \texttt{TARGET\_TEXT} represent the input/output \textbf{texts} in each language, respectively. Finally, a model-specific chat template was applied.

\begin{table}[t]
\centering
\resizebox{\columnwidth}{!}{
\begin{tabular}{@{}p{3.2cm}p{5.2cm}p{3.4cm}@{}}
\toprule
\textbf{Model family} & \textbf{Prompt / template} & \textbf{Decoding} \\
\midrule
General instruction-tuned LLMs (incl.\ ours) &
Native chat template, thinking disabled, our prompt (Fig.~\ref{fig:translation_prompt}) &
Greedy, \texttt{max\_new\_tokens}=256 \\
\midrule
TranslateGemma &
Official chat template/prompt &
Greedy, \texttt{max\_new\_tokens}=256 \\
\midrule
NLLB / AfriNLLB &
Raw source with \texttt{src\_lang} and forced target BOS token &
Greedy \\
%Beam Search, \texttt{beam}=1 \\
\midrule
Afrique-LLM &
\parbox[t]{6cm}{\texttt{[5-shot examples]\textbackslash n}\newline\texttt{\{src\} sentence: \{text\}\textbackslash n}\newline\texttt{\{tgt\} sentence:}} &
Greedy, \texttt{max\_new\_tokens}=256 \\
\bottomrule
\end{tabular}
}
\caption{Prompting and decoding configuration per baseline family.}
\label{tab:baseline_configs}
\end{table}

\subsection{Baseline Prompting and Decoding}
\label{app:baseline_configs}

We use model-appropriate prompting and decoding configurations rather than a single setup across all systems (Table~\ref{tab:baseline_configs}). General instruction-tuned LLMs, including ours, use their native chat templates with thinking disabled and the translation prompt in Figure~\ref{fig:translation_prompt}, while specialized translation models follow their corresponding prompting and decoding formats. 
The prompts did not include few-shot examples, except for AfriqueLLM models, which are base models and explicitly recommend using 5-shot examples in their instructions. In this case, the examples were selected randomly from Flores-200 \texttt{dev} split (the one used for evaluation is \texttt{devtest}). 
We evaluate the sensitivity of the main results to alternative prompting and decoding choices in Appendix~\ref{app:prompt_sensitivity}.

\subsection{Training Hyperparameters}
\label{app:hyperparams}

\paragraph{Optimization.} 
Full fine-tuning uses peak learning rate of $1.25\times10^{-5}$; fused AdamW, a linear schedule with 1\% warmup, gradient clipping at 1.0, and gradient checkpointing, with a global batch size of 256.

\paragraph{Data formatting.} 
All training samples are first converted into the translation prompt format shown in Figure~\ref{fig:translation_prompt}, and then wrapped with the model-specific chat template. Sequences exceeding 2{,}048 tokens are filtered, and remaining samples are packed via best-fit decreasing.

\paragraph{Infrastructure.} 
All experiments use PyTorch, HuggingFace Transformers, TRL, and DeepSpeed ZeRO-2 in bfloat16 on 32 NVIDIA H100 GPUs.

\begin{figure*}[t]
\centering
\includegraphics[width=\linewidth]{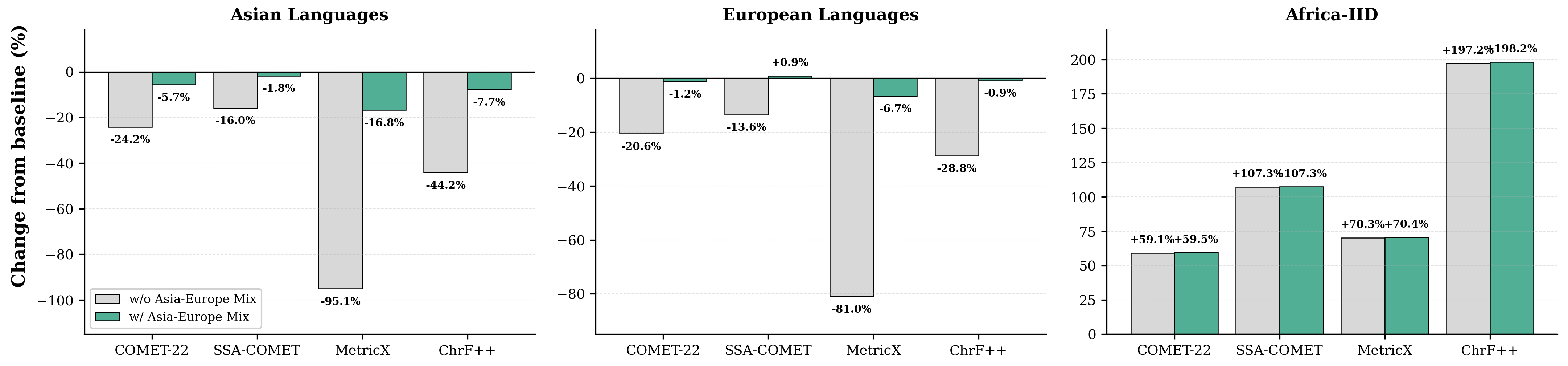}
\caption{\textbf{Effect of the Asia-Europe Mix on catastrophic forgetting and Africa-IID performance.} 
Bars show the percentage change from the Qwen3.5-2B baseline after SFT. The Asia-Europe Mix substantially reduces degradation on Asian and European language pairs while preserving nearly identical gains on the Africa-IID pairs.}
\label{fig:opus-forgetting-app}
\end{figure*}

\section{Catastrophic Forgetting Analysis}
\label{app:catastrophic-forgetting}

We further analyze whether the Asia-Europe Mix mitigates catastrophic forgetting on non-African languages without compromising African language MT performance. The mix contains bidirectional parallel data between English and non-African languages, covering both Asian\footnote{
Asian: sin, hin, ben, urd, fas, kaz, tur, zho, jpn, kor, vie, tha, ind, msa, fil.} and European\footnote{European: fra, spa, ita, por, ron, ell, lit, lav, rus, ukr, pol, slk, hrv, ces, srp, bul, deu, swe, isl, dan, nld, hun, fin, est.
} language groups.

Figure~\ref{fig:opus-forgetting-app} shows that fine-tuning without the Asia-Europe Mix substantially degrades performance on Asian and European language pairs, indicating catastrophic forgetting outside the African language training distribution. Adding the mix consistently reduces this degradation across all metrics and both language groups. For example, the MetricX-24 drop is reduced from $-95.1\%$ to $-16.8\%$ for Asian languages and from $-81.0\%$ to $-6.7\%$ for European languages.
This retention does not come at the cost of Africa-IID performance: models with and without the Asia-Europe Mix achieve nearly identical gains on African language pairs across all metrics. We therefore retain the Asia-Europe Mix in the final \modelname recipe, not to improve African language MT directly, but to preserve broader multilingual translation while maintaining the gains on the Africa-IID group.

\begin{figure*}[h]
\centering
\includegraphics[width=\linewidth]{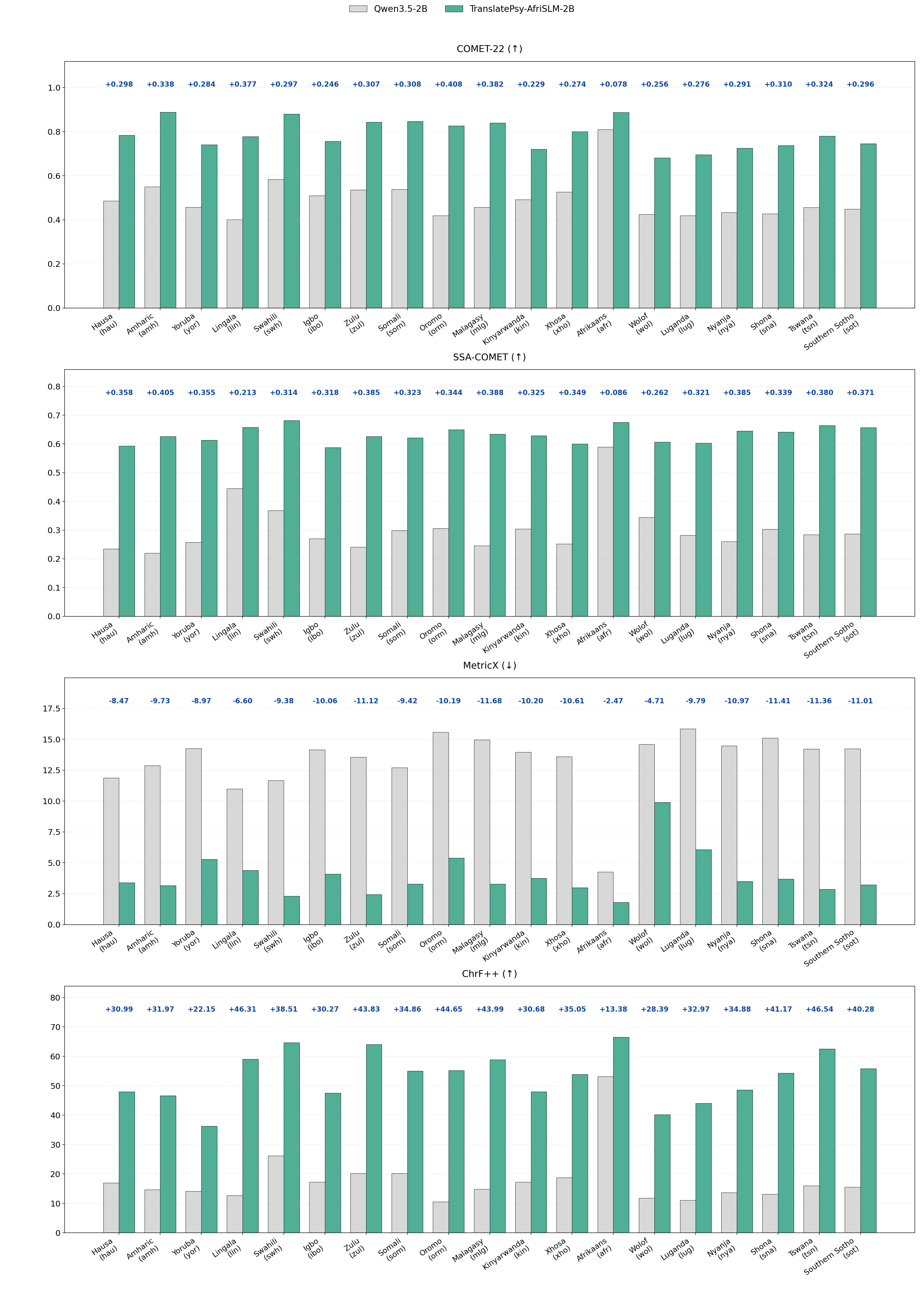}
\caption{\textbf{AFRICA-IID BOUQuET results.} Per-language scores for 19 target African languages on BOUQuET, comparing Qwen3.5-2B and \modelname-2B across four evaluation metrics. Scores are averaged over both translation directions where available. $\Delta$ = \modelname-2B $-$ baseline model scores.}
\label{fig:app-africa-iid-all-metrics}
\end{figure*}

\section{Additional Results}
\label{sec:additional_results_appendix}

\begin{table}[t]
\centering
\small
\resizebox{\columnwidth}{!}{
\begin{tabular}{lccc}
\toprule
\textbf{AFRICA-OOD lang.} & \textbf{Flores-200} & \textbf{BOUQuET} & \textbf{Smol} \\
\midrule
Nigerian Pidgin (\texttt{pcm})   &            & \checkmark & \checkmark \\
Sudanese Arabic (\texttt{apd})   &            &            & \checkmark \\
Akan (\texttt{aka})              & \checkmark & \checkmark & \checkmark \\
Tamazight (\texttt{ber})         & \checkmark & \checkmark & \checkmark \\
Kituba (\texttt{ktu})            &            &            & \checkmark \\
Bambara (\texttt{bam})           & \checkmark & \checkmark & \checkmark \\
Sepedi (\texttt{nso})            & \checkmark & \checkmark & \checkmark \\
Moor\'e (\texttt{mos})           & \checkmark & \checkmark & \checkmark \\
\midrule
\textbf{\# Langs.}               & 5          & 6          & 8          \\
\textbf{\# Directions}           & 10         & 12         & 16         \\
\bottomrule
\end{tabular}
}
\caption{\textbf{AFRICA-OOD language coverage by evaluation set.}
The number of directions counts both eng-to-xx and xx-to-eng directions.}
\label{tab:africa-ood-coverage}
\end{table}

\begin{table*}[t]
\centering
\small
\resizebox{\textwidth}{!}{
\begin{tabular}{llcccccc}
\toprule
\textbf{Data Source} & \textbf{Configuration} & \textbf{Threshold} & \textbf{Tokens} & \textbf{Threshold} & \textbf{Tokens} & \textbf{Threshold} & \textbf{Tokens} \\
\midrule
Open-Source & threshold & -0.50 & 6.57B & 0.00 & 3.79B & 0.50 & 1.76B \\
 & threshold + top-$N$ & -0.28 & 5.28B & 0.16 & 2.88B & 0.61 & 1.20B \\
 & threshold + bidirectional & -0.50 & 14.12B & 0.00 & 7.86B & 0.50 & 3.40B \\
 & threshold + top-$N$ + bidirectional & -0.28 & 9.95B & 0.16 & 5.15B & 0.61 & 2.03B \\
\midrule
Synthetic & threshold & 0.68 & 16.75B & 0.93 & 7.28B & 1.19 & 2.80B \\
 & threshold + top-$N$ & 0.71 & 12.43B & 0.93 & 4.49B & 1.18 & 1.44B \\
 & threshold + bidirectional & \textbf{0.68} & \textbf{32.37B} & 0.93 & 15.05B & 1.19 & 6.16B \\
 & threshold + top-$N$ + bidirectional & 0.71 & 23.62B & 0.93 & 9.09B & 1.18 & 3.32B \\
\midrule
Open-Source + Synthetic & threshold & 0.41 & 23.32B & 0.67 & 11.08B & 0.97 & 4.56B \\
 & threshold + top-$N$ & 0.55 & 17.71B & 0.78 & 7.37B & 1.07 & 2.64B \\
 & threshold + bidirectional & 0.41 & 46.49B & 0.67 & 22.92B & 0.97 & 9.56B \\
 & threshold + top-$N$ + bidirectional & 0.55 & 33.57B & 0.78 & 14.24B & 1.07 & 5.35B \\
\bottomrule
\end{tabular}
}
\caption{\textbf{Threshold values and training token counts for each data source, configuration, and filtering level} (for the datasets plotted in Figure~\ref{fig:scaling_pareto_plot}). Since the threshold may vary depending on the data mix and language direction, and topN changes the effective threshold, the value shown is an average weighted by the number of example pairs at each actual threshold. The bolded configuration indicates the \modelname data mix used for final training.}
\label{tab:threshold_training_tokens}
\end{table*}

\subsection{Per-Language Performance Analysis}
\label{app:per-language}

\paragraph{Africa-IID}
Figure~\ref{fig:app-africa-iid-all-metrics} shows per-language results on the Africa-IID BOUQuET benchmark, averaging scores over both eng-xx and xx-eng directions. \modelname-2B consistently improves over the Qwen3.5-2B baseline across the 19 in-distribution languages and all four evaluation metrics. The largest gains appear for low-baseline languages such as Oromo, Malagasy, Lingala, Tswana, and Zulu, while Afrikaans shows smaller gains due to its stronger baseline.

\paragraph{Africa-OOD}
Figure~\ref{fig:app-africa-ood-all-metrics} reports results for Africa-OOD languages held out from fine-tuning. Because Flores-200, BOUQuET, and Smol cover different subsets of OOD languages, scores are averaged over both directions and available evaluation sets, as summarized in Table~\ref{tab:africa-ood-coverage}. Although the OOD languages are held out from fine-tuning, \modelname-2B still improves overall performance, suggesting meaningful transfer beyond the languages seen during post-training. The strongest gains appear for Sepedi, Bambara, Akan, and Moor\'e, while improvements are smaller and more heterogeneous than in the IID setting, with some metric-specific regressions for Nigerian Pidgin, Sudanese Arabic, and Tamazight.

\begin{table}[t]
\centering
\small
\resizebox{\columnwidth}{!}{
\begin{tabular}{lcccc}
\toprule
\textbf{Model} & C22 ($\uparrow$) & SSA ($\uparrow$) & MX ($\downarrow$) & chrF++ ($\uparrow$) \\
\midrule
\multicolumn{5}{c}{\textit{\textbf{General-purpose LLMs}}} \\
\midrule
Apertus-8B & 0.4624 & 0.2737 & 14.973 & 11.141 \\
\midrule
Qwen3.5-0.8B & 0.3716 & 0.3089 & 16.627 & 5.285 \\
Qwen3.5-2B & 0.4086 & 0.2925 & 16.523 & 7.418 \\
Qwen3.5-4B & 0.4805 & 0.3079 & 15.035 & 10.871 \\
Qwen3.5-27B & 0.6883 & 0.4117 & 8.050 & 26.303 \\
Qwen3.5-122B-A10B & 0.7214 & 0.4602 & 6.525 & 29.842 \\
\midrule
\multicolumn{5}{c}{\textit{\textbf{Dedicated translation models}}} \\
\midrule
AfriNLLB-600M & 0.7683 & 0.4974 & 5.777 & 34.715 \\
NLLB-1.3B & 0.7555 & 0.4821 & 6.299 & 32.758 \\
NLLB-3.3B & 0.7582 & 0.4907 & 5.987 & 33.317 \\
\midrule
Hunyuan-MT-7B & 0.4187 & 0.4163 & 12.812 & 9.077 \\
\midrule
TranslateGemma-4B & 0.4691 & 0.2650 & 12.688 & 10.384 \\
TranslateGemma-27B & 0.7590 & 0.4878 & 6.058 & 31.922 \\
\midrule
\multicolumn{5}{c}{\textit{\textbf{African language-specialized models}}} \\
\midrule
AfriqueLlama-8B & 0.7035 & 0.4674 & 6.315 & 28.910 \\
AfriqueGemma-12B & 0.7317 & 0.5009 & 5.360 & 33.213 \\
AfriqueQwen-14B & 0.7278 & 0.4876 & 5.604 & 31.196 \\
\midrule
\multicolumn{5}{c}{\textit{\textbf{Ours}}} \\
\midrule
\modelname-0.8B & 0.7276 & 0.4810 & 6.990 & 28.576 \\
\modelname-2B & 0.7694 & 0.5271 & 5.333 & 34.039 \\
\modelname-4B & \textbf{0.7855} & \textbf{0.5503} & \textbf{4.617} & \textbf{36.749} \\
\bottomrule
\end{tabular}
}
\vspace{0.5em}
\caption{\textbf{African-to-African language translation results on BOUQuET}, averaged over all 20 directions among the five highest-resource hub languages (swh, hau, yor, zul, amh). None of these xx$\to$yy directions appear in training (zero-shot). Best per column in \textbf{bold}.}
\label{tab:afr2afr_bouquet}
\end{table}

\begin{table}[t]
\centering
\small
\resizebox{\columnwidth}{!}{
\begin{tabular}{lccccc}
\toprule
\textbf{src\,$\downarrow$ / tgt\,$\rightarrow$} & \textbf{swh} & \textbf{hau} & \textbf{yor} & \textbf{zul} & \textbf{amh} \\
\midrule
\textbf{swh} & -- & 0.5746 & 0.5994 & 0.6123 & 0.6096 \\
\textbf{hau} & 0.5856 & -- & 0.5173 & 0.5090 & 0.5211 \\
\textbf{yor} & 0.5633 & 0.4483 & -- & 0.4665 & 0.4896 \\
\textbf{zul} & 0.6346 & 0.5148 & 0.5457 & -- & 0.5507 \\
\textbf{amh} & 0.6157 & 0.5192 & 0.5843 & 0.5450 & -- \\
\bottomrule
\end{tabular}
}
\vspace{0.5em}
\caption{\textbf{Per-direction SSA-COMET ($\uparrow$) for \modelname-4B on BOUQuET African-to-African language translation.} Rows are source, columns are target languages; the model is never trained on any xx$\to$yy pair.}
\label{tab:afr2afr_directions}
\end{table}

\subsection{Zero-Shot African-to-African Translation} 

\label{sec:afr2afr} Our training data consists solely of English$\leftrightarrow$African language (en$\leftrightarrow$xx) synthetic pairs; thus, direct African-to-African language (xx$\to$yy) translation directions are never observed during training. We evaluate whether the improvements obtained from our training pipeline transfer to these unseen directions, an important but under-served translation setting. 

We evaluate on BOUQuET over all 20 ordered directions ($5\times4$) among five high-resource African hub languages: Swahili (swh), Hausa (hau), Yoruba (yor), Zulu (zul), and Amharic (amh). Table~\ref{tab:afr2afr_bouquet} reports results averaged across the 20 directions, while Table~\ref{tab:afr2afr_directions} provides the per-direction SSA-COMET scores for \modelname-4B. 

Despite having no direct African-to-African language training pairs, \modelname-4B achieves the best aggregate performance on all four metrics, outperforming the strongest baselines across general-purpose LLMs, dedicated translation models, and African language-specialized models. These results suggest that the gains from our training pipeline transfer beyond the en$\leftrightarrow$xx directions seen during training to unseen African-to-African language translation directions.

\begin{table}[t]
\centering
\resizebox{\columnwidth}{!}{
\begin{tabular}{@{}llcccc@{}}
\toprule
\textbf{Model} & \textbf{Variant} & \textbf{COMET-22} & \textbf{SSA-COMET}
& \textbf{MetricX}$\downarrow$ & \textbf{chrF++} \\
\midrule
\multirow{3}{*}{Qwen3.5-27B}
 & Default    & 78.0 & 53.8 & 5.52 & 44.7 \\
 & Minimal    & 78.0 & 53.7 & 5.62 & 45.6 \\
 & Structured & 78.1 & 53.8 & 5.53 & 45.3 \\
\midrule
\multirow{3}{*}{TranslateGemma-27B}
 & Official, greedy         & 80.0 & 57.5 & 4.54 & 46.5 \\
 & Official, sampling       & 79.8 & 56.9 & 4.68 & 46.0 \\
 & Generic, no chat template & 80.0 & 57.6 & 4.49 & 47.4 \\
\bottomrule
\end{tabular}
}
\caption{Sensitivity to prompt/decoding variants on 10 Flores-200 SSA pairs. Main-table ordering is unchanged across all settings.}
\label{tab:sensitivity}
\end{table}

\begin{table}[t]
\centering
\small
%\begin{tabular}{lcc@{\hspace{4pt}}c@{\hspace{3pt}}c}
\begin{tabular}{lcccc}
\toprule
Quality estimator & C22 & SSA & \ MX\ & chrF++ \\
\midrule
\multicolumn{5}{c}{Results on BOUQuET test set.} \\
\midrule
Random & 0.761 & 0.624 & 4.21 & 50.0 \\
z-score & 0.765 & 0.643 & 3.80 & 50.1 \\
SSA-COMET & \textbf{0.766} & \textbf{0.647} & 3.96 & \textbf{50.5} \\
AfriCOMET & 0.763 & 0.637 & 4.03 & 49.8 \\
MetricX & 0.760 & 0.632 & \textbf{3.62} & 49.6 \\
\midrule
\multicolumn{5}{c}{Results on Flores-200 test set.} \\
\midrule
Random & 0.727 & 0.590 & 5.22 & 45.1 \\
z-score & \textbf{0.737} & 0.617 & 4.52 & 45.7 \\
SSA-COMET & 0.736 & \textbf{0.620} & 4.73 & \textbf{45.9} \\
AfriCOMET & 0.735 & 0.609 & 4.87 & 45.3 \\
MetricX & 0.730 & 0.605 & \textbf{4.29} & 45.2 \\
\midrule
\multicolumn{5}{c}{Results on Smol test set.} \\
\midrule
Random & 0.646 & 0.476 & 9.59 & 27.8 \\
z-score & \textbf{0.655} & 0.503 & 8.88 & \textbf{28.0} \\
SSA-COMET & 0.652 & \textbf{0.505} & 9.22 & \textbf{28.0} \\
AfriCOMET & 0.653 & 0.495 & 9.26 & 27.8 \\
MetricX & 0.650 & 0.491 & \textbf{8.58} & 27.9 \\
\bottomrule
\end{tabular}
\caption{\textbf{Translation quality with individual versus unified QE} for English--African language translation.}
\label{tab:indiv_qe_metrics}
\end{table}

\begin{table}[h]
\centering
\small
\resizebox{\columnwidth}{!}{
\begin{tabular}{lcccccccc}
\toprule
 & \multicolumn{2}{c}{C22} & \multicolumn{2}{c}{SSA} & \multicolumn{2}{c}{MX} & \multicolumn{2}{c}{chrF++} \\
\cmidrule(lr){2-3} \cmidrule(lr){4-5} \cmidrule(lr){6-7} \cmidrule(lr){8-9}
\parbox{2.2cm}{\centering Comparison\\($\bar{z}$ vs.\ metric)} & $\Delta$ & $p$ & $\Delta$ & $p$ & $\Delta$ & $p$ & $\Delta$ & $p$ \\
\midrule
\multicolumn{9}{c}{BOUQuET} \\
\midrule
$\bar{z}$ vs.\ SSA-COMET & \textcolor{red}{-0.001} & 0.004 & \textcolor{red}{-0.004} & $<$0.001 & \textcolor{red}{-0.16} & $<$0.001 & \textcolor{red}{-0.5} & $<$0.001 \\
$\bar{z}$ vs.\ AfriCOMET & \textcolor{blue}{+0.002} & $<$0.001 & \textcolor{blue}{+0.006} & $<$0.001 & \textcolor{red}{-0.22} & $<$0.001 & \textcolor{blue}{+0.3} & $<$0.001 \\
$\bar{z}$ vs.\ MetricX & \textcolor{blue}{+0.005} & $<$0.001 & \textcolor{blue}{+0.011} & $<$0.001 & \textcolor{red}{+0.19} & $<$0.001 & \textcolor{blue}{+0.4} & $<$0.001 \\
\midrule
\multicolumn{9}{c}{Flores-200} \\
\midrule
$\bar{z}$ vs.\ SSA-COMET & 0.000 & \textbf{0.113} & \textcolor{red}{-0.003} & $<$0.001 & \textcolor{red}{-0.22} & $<$0.001 & \textcolor{red}{-0.2} & $<$0.001 \\
$\bar{z}$ vs.\ AfriCOMET & \textcolor{blue}{+0.002} & $<$0.001 & \textcolor{blue}{+0.008} & $<$0.001 & \textcolor{red}{-0.36} & $<$0.001 & \textcolor{blue}{+0.4} & $<$0.001 \\
$\bar{z}$ vs.\ MetricX & \textcolor{blue}{+0.006} & $<$0.001 & \textcolor{blue}{+0.013} & $<$0.001 & \textcolor{red}{+0.23} & $<$0.001 & \textcolor{blue}{+0.4} & $<$0.001 \\
\midrule
\multicolumn{9}{c}{Smol} \\
\midrule
$\bar{z}$ vs.\ SSA-COMET & \textcolor{blue}{+0.003} & $<$0.001 & \textcolor{red}{-0.002} & $<$0.001 & \textcolor{red}{-0.34} & $<$0.001 & 0.0 & \textbf{0.348} \\
$\bar{z}$ vs.\ AfriCOMET & \textcolor{blue}{+0.001} & 0.005 & \textcolor{blue}{+0.009} & $<$0.001 & \textcolor{red}{-0.37} & $<$0.001 & \textcolor{blue}{+0.1} & $<$0.001 \\
$\bar{z}$ vs.\ MetricX & \textcolor{blue}{+0.004} & $<$0.001 & \textcolor{blue}{+0.012} & $<$0.001 & \textcolor{red}{+0.30} & $<$0.001 & \textcolor{blue}{+0.1} & 0.034 \\
\bottomrule
\end{tabular}
}
\caption{\textbf{Paired significance tests for the QE-metric ablation} ($B=10{,}000$). $\Delta$ is $\bar{z}$ minus the individual filter. Blue/red denotes better/worse $\bar{z}$ performance (lower is better for MetricX); bold denotes $p\ge0.05$.}
\label{tab:qe_metrics_significance}
\end{table}

\section{Additional Evaluation and Robustness}

\subsection{Prompt and Decoding Sensitivity} \label{app:prompt_sensitivity} To assess whether our baseline comparisons are sensitive to prompting or decoding choices, we evaluate three variants each for Qwen3.5-27B and TranslateGemma-27B on 10 Flores-200 SSA language pairs (Table~\ref{tab:sensitivity}). For Qwen3.5-27B, we compare the default prompt with minimal and structured variants. For TranslateGemma-27B, we compare the official greedy setup with sampling and a generic plain-completion prompt without the chat template. Across these variants, the maximum differences are 0.2 COMET-22, 0.7 SSA-COMET, 0.19 MetricX, and 1.4 chrF++. Importantly, the main-table ordering remains unchanged, and TranslateGemma-27B remains among the strongest baselines across all tested settings. These results indicate that our baseline comparisons are robust to the prompting and decoding variations considered here.

\subsection{Statistical Significance (Bootstrap)}
\label{sec:bootstrap_significance}

We assess statistical significance using paired bootstrap resampling over sentence pairs. For each model pair, dataset, and metric, we align common translation directions, pair sentence-level scores, and draw $B=10{,}000$ bootstrap samples with replacement. We report $\Delta$ as the mean score difference and compute $p$ as the fraction of bootstrap samples that do not support the observed direction of improvement. For MetricX, lower scores are better; for all other metrics, higher scores are better.

Table~\ref{tab:significance} shows that the main rankings are highly stable across benchmarks and metrics. \modelname-0.8B substantially outperforms the much larger general-purpose Qwen3.5-122B on SSA-COMET, MetricX, and chrF++, and remains competitive on COMET-22. It also consistently surpasses TranslateGemma-27B and AfriqueGemma-12B across all reported settings. Against NLLB, \modelname-0.8B is competitive but mixed, whereas \modelname-2B significantly outperforms both NLLB-1.3B and NLLB-3.3B across every benchmark--metric combination. Within the \modelname family, scaling from 0.8B to 2B and from 2B to 4B yields monotonic and statistically significant gains. Overall, the bootstrap results confirm that the observed improvements are stable under paired sentence-level resampling rather than being driven by evaluation noise.

\begin{table*}[p]
\centering
\small
\resizebox{\textwidth}{!}{
\begin{tabular}{llcccccc}
\toprule
 & & \multicolumn{2}{c}{\textbf{FLORES-200}} & \multicolumn{2}{c}{\textbf{BOUQuET}} & \multicolumn{2}{c}{\textbf{SMOL}} \\
\cmidrule(lr){3-4} \cmidrule(lr){5-6} \cmidrule(lr){7-8}
\textbf{Comparison} & \textbf{Metric} & $\Delta$ & $p$ & $\Delta$ & $p$ & $\Delta$ & $p$ \\
\midrule
\multicolumn{8}{l}{\textit{\textbf{vs.\ General-purpose LLM}}} \\
\midrule
\modelname-0.8B vs.\ Qwen3.5-122B & COMET-22 & \textcolor{red}{-0.0001} & 0.452 & \textcolor{blue}{+0.0135} & $<$0.001 & \textcolor{blue}{+0.0017} & $<$0.001 \\
\modelname-0.8B vs.\ Qwen3.5-122B & SSA-COMET & \textcolor{blue}{+0.0413} & $<$0.001 & \textcolor{blue}{+0.0379} & $<$0.001 & \textcolor{blue}{+0.0286} & $<$0.001 \\
\modelname-0.8B vs.\ Qwen3.5-122B & MetricX & \textcolor{blue}{-0.548} & $<$0.001 & \textcolor{blue}{-0.723} & $<$0.001 & \textcolor{blue}{-0.506} & $<$0.001 \\
\modelname-0.8B vs.\ Qwen3.5-122B & chrF++ & \textcolor{blue}{+3.30} & $<$0.001 & \textcolor{blue}{+5.33} & $<$0.001 & \textcolor{blue}{+1.29} & $<$0.001 \\
\modelname-0.8B vs.\ Qwen3.5-122B & spBLEU & \textcolor{blue}{+3.95} & $<$0.001 & \textcolor{blue}{+7.38} & $<$0.001 & \textcolor{blue}{+1.88} & $<$0.001 \\
\midrule
\multicolumn{8}{l}{\textit{\textbf{vs.\ Dedicated Translation}}} \\
\midrule
\modelname-0.8B vs.\ NLLB-1.3B & COMET-22 & \textcolor{red}{-0.0039} & $<$0.001 & \textcolor{blue}{+0.0011} & 0.017 & \textcolor{blue}{+0.0021} & $<$0.001 \\
\modelname-0.8B vs.\ NLLB-1.3B & SSA-COMET & \textcolor{blue}{+0.0066} & $<$0.001 & \textcolor{blue}{+0.0086} & $<$0.001 & \textcolor{blue}{+0.0122} & $<$0.001 \\
\modelname-0.8B vs.\ NLLB-1.3B & MetricX & \textcolor{red}{+0.038} & 0.002 & \textcolor{blue}{-0.154} & $<$0.001 & \textcolor{blue}{-0.242} & $<$0.001 \\
\modelname-0.8B vs.\ NLLB-1.3B & chrF++ & \textcolor{red}{-0.40} & $<$0.001 & \textcolor{blue}{+0.48} & $<$0.001 & \textcolor{red}{-0.01} & 0.395 \\
\modelname-0.8B vs.\ NLLB-1.3B & spBLEU & \textcolor{blue}{+0.16} & 0.004 & \textcolor{blue}{+1.57} & $<$0.001 & \textcolor{blue}{+0.24} & $<$0.001 \\
\midrule
\modelname-0.8B vs.\ NLLB-3.3B & COMET-22 & \textcolor{red}{-0.0093} & $<$0.001 & \textcolor{blue}{+0.0064} & $<$0.001 & \textcolor{blue}{+0.0011} & 0.002 \\
\modelname-0.8B vs.\ NLLB-3.3B & SSA-COMET & \textcolor{blue}{+0.0000} & 0.483 & \textcolor{blue}{+0.0133} & $<$0.001 & \textcolor{blue}{+0.0124} & $<$0.001 \\
\modelname-0.8B vs.\ NLLB-3.3B & MetricX & \textcolor{red}{+0.284} & $<$0.001 & \textcolor{blue}{-0.007} & 0.314 & \textcolor{red}{+0.004} & 0.385 \\
\modelname-0.8B vs.\ NLLB-3.3B & chrF++ & \textcolor{red}{-1.45} & $<$0.001 & \textcolor{blue}{+0.12} & 0.101 & \textcolor{red}{-0.19} & $<$0.001 \\
\modelname-0.8B vs.\ NLLB-3.3B & spBLEU & \textcolor{red}{-0.93} & $<$0.001 & \textcolor{red}{-0.30} & 0.005 & \textcolor{red}{-0.10} & 0.027 \\
\midrule
\modelname-2B vs.\ NLLB-1.3B & COMET-22 & \textcolor{blue}{+0.0081} & $<$0.001 & \textcolor{blue}{+0.0119} & $<$0.001 & \textcolor{blue}{+0.0091} & $<$0.001 \\
\modelname-2B vs.\ NLLB-1.3B & SSA-COMET & \textcolor{blue}{+0.0193} & $<$0.001 & \textcolor{blue}{+0.0189} & $<$0.001 & \textcolor{blue}{+0.0227} & $<$0.001 \\
\modelname-2B vs.\ NLLB-1.3B & MetricX & \textcolor{blue}{-0.522} & $<$0.001 & \textcolor{blue}{-0.565} & $<$0.001 & \textcolor{blue}{-0.624} & $<$0.001 \\
\modelname-2B vs.\ NLLB-1.3B & chrF++ & \textcolor{blue}{+1.28} & $<$0.001 & \textcolor{blue}{+2.42} & $<$0.001 & \textcolor{blue}{+0.75} & $<$0.001 \\
\modelname-2B vs.\ NLLB-1.3B & spBLEU & \textcolor{blue}{+2.00} & $<$0.001 & \textcolor{blue}{+3.76} & $<$0.001 & \textcolor{blue}{+0.96} & $<$0.001 \\
\midrule
\modelname-2B vs.\ NLLB-3.3B & COMET-22 & \textcolor{blue}{+0.0027} & $<$0.001 & \textcolor{blue}{+0.0172} & $<$0.001 & \textcolor{blue}{+0.0081} & $<$0.001 \\
\modelname-2B vs.\ NLLB-3.3B & SSA-COMET & \textcolor{blue}{+0.0126} & $<$0.001 & \textcolor{blue}{+0.0237} & $<$0.001 & \textcolor{blue}{+0.0229} & $<$0.001 \\
\modelname-2B vs.\ NLLB-3.3B & MetricX & \textcolor{blue}{-0.276} & $<$0.001 & \textcolor{blue}{-0.419} & $<$0.001 & \textcolor{blue}{-0.378} & $<$0.001 \\
\modelname-2B vs.\ NLLB-3.3B & chrF++ & \textcolor{blue}{+0.24} & $<$0.001 & \textcolor{blue}{+2.06} & $<$0.001 & \textcolor{blue}{+0.57} & $<$0.001 \\
\modelname-2B vs.\ NLLB-3.3B & spBLEU & \textcolor{blue}{+0.90} & $<$0.001 & \textcolor{blue}{+1.89} & $<$0.001 & \textcolor{blue}{+0.62} & $<$0.001 \\
\midrule
\modelname-0.8B vs.\ TranslateGemma-27B & COMET-22 & \textcolor{blue}{+0.0097} & $<$0.001 & \textcolor{blue}{+0.0272} & $<$0.001 & \textcolor{blue}{+0.0018} & $<$0.001 \\
\modelname-0.8B vs.\ TranslateGemma-27B & SSA-COMET & \textcolor{blue}{+0.0489} & $<$0.001 & \textcolor{blue}{+0.0547} & $<$0.001 & \textcolor{blue}{+0.0365} & $<$0.001 \\
\modelname-0.8B vs.\ TranslateGemma-27B & MetricX & \textcolor{blue}{-0.945} & $<$0.001 & \textcolor{blue}{-1.190} & $<$0.001 & \textcolor{blue}{-0.643} & $<$0.001 \\
\modelname-0.8B vs.\ TranslateGemma-27B & chrF++ & \textcolor{blue}{+5.08} & $<$0.001 & \textcolor{blue}{+7.79} & $<$0.001 & \textcolor{blue}{+2.33} & $<$0.001 \\
\modelname-0.8B vs.\ TranslateGemma-27B & spBLEU & \textcolor{blue}{+5.30} & $<$0.001 & \textcolor{blue}{+7.85} & $<$0.001 & \textcolor{blue}{+2.53} & $<$0.001 \\
\midrule
\multicolumn{8}{l}{\textit{\textbf{vs.\ African-Specialized}}} \\
\midrule
\modelname-0.8B vs.\ AfriqueGemma-12B & COMET-22 & \textcolor{blue}{+0.0301} & $<$0.001 & \textcolor{blue}{+0.0204} & $<$0.001 & \textcolor{blue}{+0.0212} & $<$0.001 \\
\modelname-0.8B vs.\ AfriqueGemma-12B & SSA-COMET & \textcolor{blue}{+0.0414} & $<$0.001 & \textcolor{blue}{+0.0306} & $<$0.001 & \textcolor{blue}{+0.0284} & $<$0.001 \\
\modelname-0.8B vs.\ AfriqueGemma-12B & MetricX & \textcolor{blue}{-0.664} & $<$0.001 & \textcolor{blue}{-0.531} & $<$0.001 & \textcolor{blue}{-0.444} & $<$0.001 \\
\modelname-0.8B vs.\ AfriqueGemma-12B & chrF++ & \textcolor{blue}{+3.57} & $<$0.001 & \textcolor{blue}{+3.19} & $<$0.001 & \textcolor{blue}{+2.03} & $<$0.001 \\
\modelname-0.8B vs.\ AfriqueGemma-12B & spBLEU & \textcolor{blue}{+1.43} & $<$0.001 & \textcolor{blue}{+1.69} & $<$0.001 & \textcolor{blue}{+0.94} & $<$0.001 \\
\midrule
\multicolumn{8}{l}{\textit{\textbf{Scaling Effect}}} \\
\midrule
\modelname-4B vs.\ \modelname-2B & COMET-22 & \textcolor{blue}{+0.0077} & $<$0.001 & \textcolor{blue}{+0.0080} & $<$0.001 & \textcolor{blue}{+0.0059} & $<$0.001 \\
\modelname-4B vs.\ \modelname-2B & SSA-COMET & \textcolor{blue}{+0.0073} & $<$0.001 & \textcolor{blue}{+0.0071} & $<$0.001 & \textcolor{blue}{+0.0063} & $<$0.001 \\
\modelname-4B vs.\ \modelname-2B & MetricX & \textcolor{blue}{-0.320} & $<$0.001 & \textcolor{blue}{-0.287} & $<$0.001 & \textcolor{blue}{-0.260} & $<$0.001 \\
\modelname-4B vs.\ \modelname-2B & chrF++ & \textcolor{blue}{+1.14} & $<$0.001 & \textcolor{blue}{+1.33} & $<$0.001 & \textcolor{blue}{+0.65} & $<$0.001 \\
\modelname-4B vs.\ \modelname-2B & spBLEU & \textcolor{blue}{+1.32} & $<$0.001 & \textcolor{blue}{+1.63} & $<$0.001 & \textcolor{blue}{+0.62} & $<$0.001 \\
\midrule
\modelname-2B vs.\ \modelname-0.8B & COMET-22 & \textcolor{blue}{+0.0120} & $<$0.001 & \textcolor{blue}{+0.0116} & $<$0.001 & \textcolor{blue}{+0.0085} & $<$0.001 \\
\modelname-2B vs.\ \modelname-0.8B & SSA-COMET & \textcolor{blue}{+0.0126} & $<$0.001 & \textcolor{blue}{+0.0163} & $<$0.001 & \textcolor{blue}{+0.0152} & $<$0.001 \\
\modelname-2B vs.\ \modelname-0.8B & MetricX & \textcolor{blue}{-0.560} & $<$0.001 & \textcolor{blue}{-0.552} & $<$0.001 & \textcolor{blue}{-0.441} & $<$0.001 \\
\modelname-2B vs.\ \modelname-0.8B & chrF++ & \textcolor{blue}{+1.68} & $<$0.001 & \textcolor{blue}{+1.96} & $<$0.001 & \textcolor{blue}{+0.83} & $<$0.001 \\
\modelname-2B vs.\ \modelname-0.8B & spBLEU & \textcolor{blue}{+1.83} & $<$0.001 & \textcolor{blue}{+2.17} & $<$0.001 & \textcolor{blue}{+0.72} & $<$0.001 \\
\midrule
\modelname-4B vs.\ \modelname-0.8B & COMET-22 & \textcolor{blue}{+0.0197} & $<$0.001 & \textcolor{blue}{+0.0196} & $<$0.001 & \textcolor{blue}{+0.0143} & $<$0.001 \\
\modelname-4B vs.\ \modelname-0.8B & SSA-COMET & \textcolor{blue}{+0.0199} & $<$0.001 & \textcolor{blue}{+0.0234} & $<$0.001 & \textcolor{blue}{+0.0214} & $<$0.001 \\
\modelname-4B vs.\ \modelname-0.8B & MetricX & \textcolor{blue}{-0.880} & $<$0.001 & \textcolor{blue}{-0.840} & $<$0.001 & \textcolor{blue}{-0.701} & $<$0.001 \\
\modelname-4B vs.\ \modelname-0.8B & chrF++ & \textcolor{blue}{+2.82} & $<$0.001 & \textcolor{blue}{+3.29} & $<$0.001 & \textcolor{blue}{+1.49} & $<$0.001 \\
\modelname-4B vs.\ \modelname-0.8B & spBLEU & \textcolor{blue}{+3.16} & $<$0.001 & \textcolor{blue}{+3.80} & $<$0.001 & \textcolor{blue}{+1.35} & $<$0.001 \\
\bottomrule
\end{tabular}
} 
\vspace{0.3em}
\caption{\textbf{Paired bootstrap significance tests} ($B=10{,}000$). $\Delta$ = score difference (ours $-$ baseline). Baseline comparisons use 0.8B models. Blue $\Delta$ = ours better; red = ours worse (MetricX: lower is better, negative $\Delta$ is blue).}
\label{tab:significance}
\end{table*}

\subsection{Robustness of QE-Metric Selection}
\label{sec:qe_metric_selection}

We apply the same paired bootstrap procedure to the QE-filter ablation in §\ref{sec:data_estimators_results} and Table~\ref{tab:indiv_qe_metrics}, comparing the unified $\bar{z}$ filter against each individual estimator (SSA-COMET, AfriCOMET, and MetricX) on BOUQuET, Flores-200, and Smol %in both eng$\to$xx and xx$\to$eng directions 
(Table~\ref{tab:qe_metrics_significance}). %--\ref{tab:qe_metrics_significance_smol}). 
Of the 36 comparisons, 34 (94.4\%) reach $p<0.05$ (BOUQuET 12/12, Flores-200 and Smol 11/12). However, the absolute score differences remain small: $|\Delta|\le0.006$ on COMET-22, $\le0.013$ on SSA-COMET, $\le0.37$ on MetricX, and $\le0.5$ on chrF++. With $B=10{,}000$ and thousands of sentence pairs, even small differences can be statistically detectable; the tests therefore primarily indicate the consistency of the observed differences across resamples.

The main finding is robustness rather than uniform dominance. Single-metric filters generally perform best on their corresponding evaluator but can lose ground on others. In contrast, $\bar{z}$ remains consistently competitive across metrics, test sets, and translation directions without a large deficit on any single metric. We therefore select $\bar{z}$ as a robust choice under heterogeneous evaluation rather than as a uniformly superior estimator.

%\vspace{-0.5em}

%\subsection{Evaluation Metric Circularity}
%\label{sec:metric_circularity}

\subsection{LLM-as-a-Judge Evaluation}
\label{sec:llm_as_a_judge}
\begin{figure*}[t]
\centering
\vspace{-2em}

\includegraphics[
    width=\linewidth,
    height=0.4\textheight,
    keepaspectratio
]{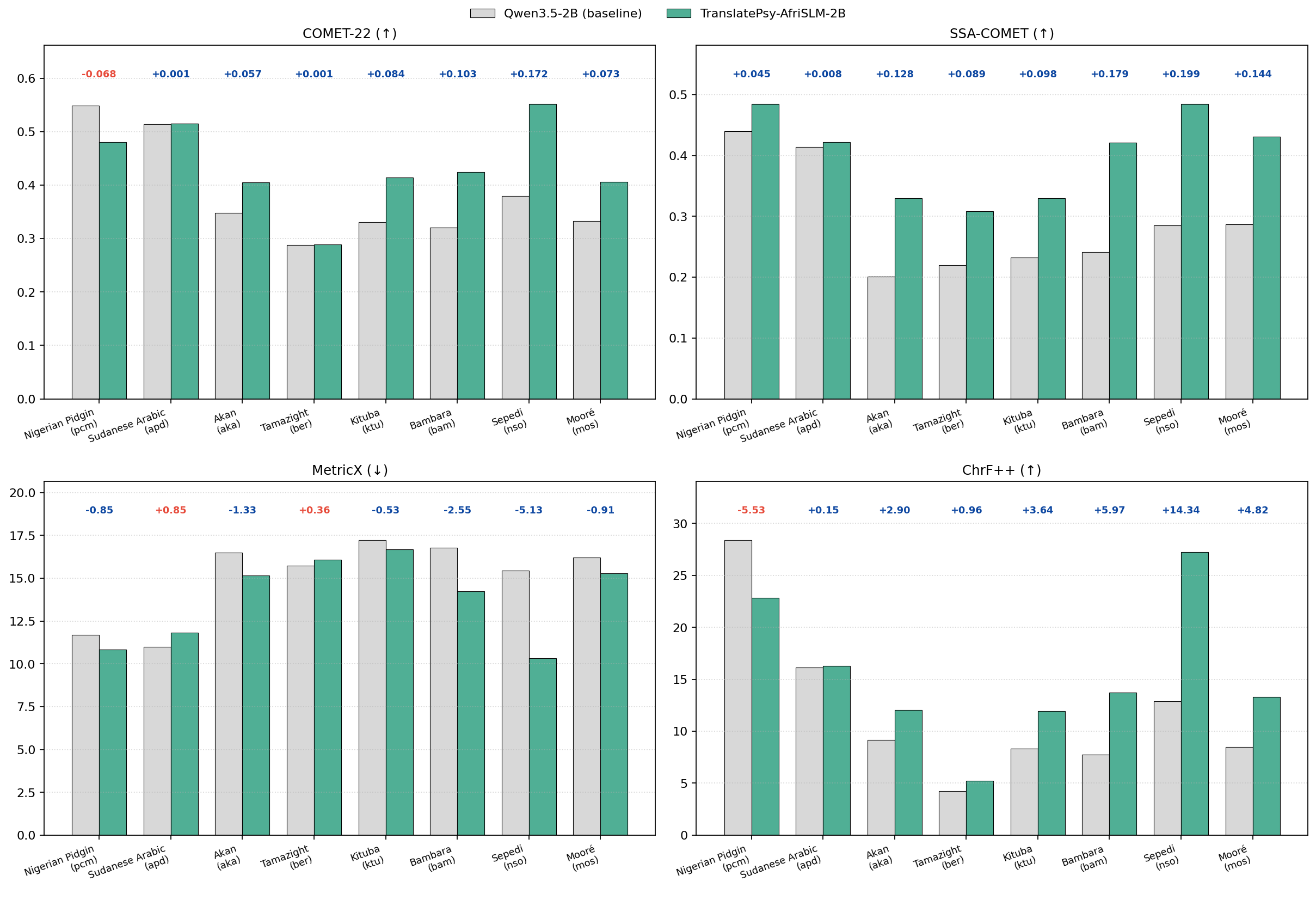}

\vspace{-0.4em}
\caption{\textbf{AFRICA-OOD results across all evaluation metrics.}
Per-language scores for held-out African languages, comparing Qwen3.5-2B and \modelname-2B.
Scores are averaged over both translation directions and over the available evaluation sets among Flores-200, BOUQuET, and Smol; language coverage differs by evaluation set as shown in Table~\ref{tab:africa-ood-coverage}.
$\Delta$ = \modelname-2B $-$ baseline model scores.}
\label{fig:app-africa-ood-all-metrics}

\vspace{2.5em}

\includegraphics[
    width=\linewidth,
    height=0.24\textheight,
    keepaspectratio
]{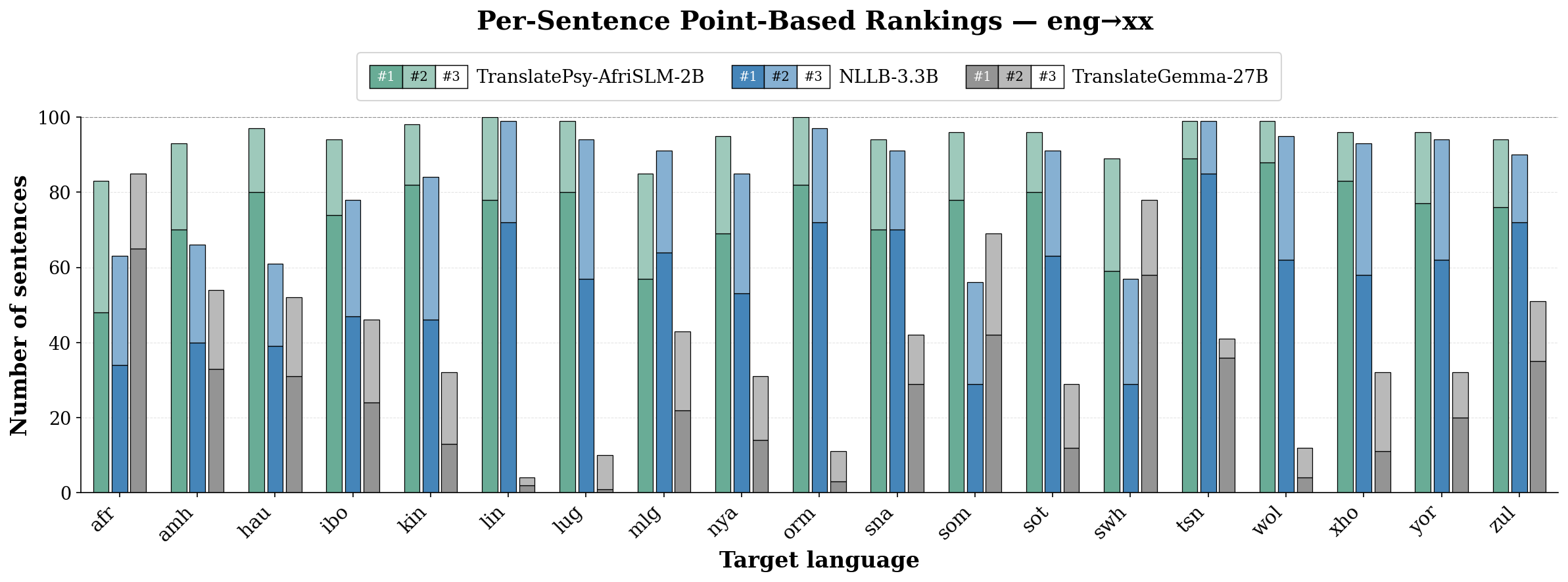}

\vspace{0.8em}

\includegraphics[
    width=\linewidth,
    height=0.24\textheight,
    keepaspectratio
]{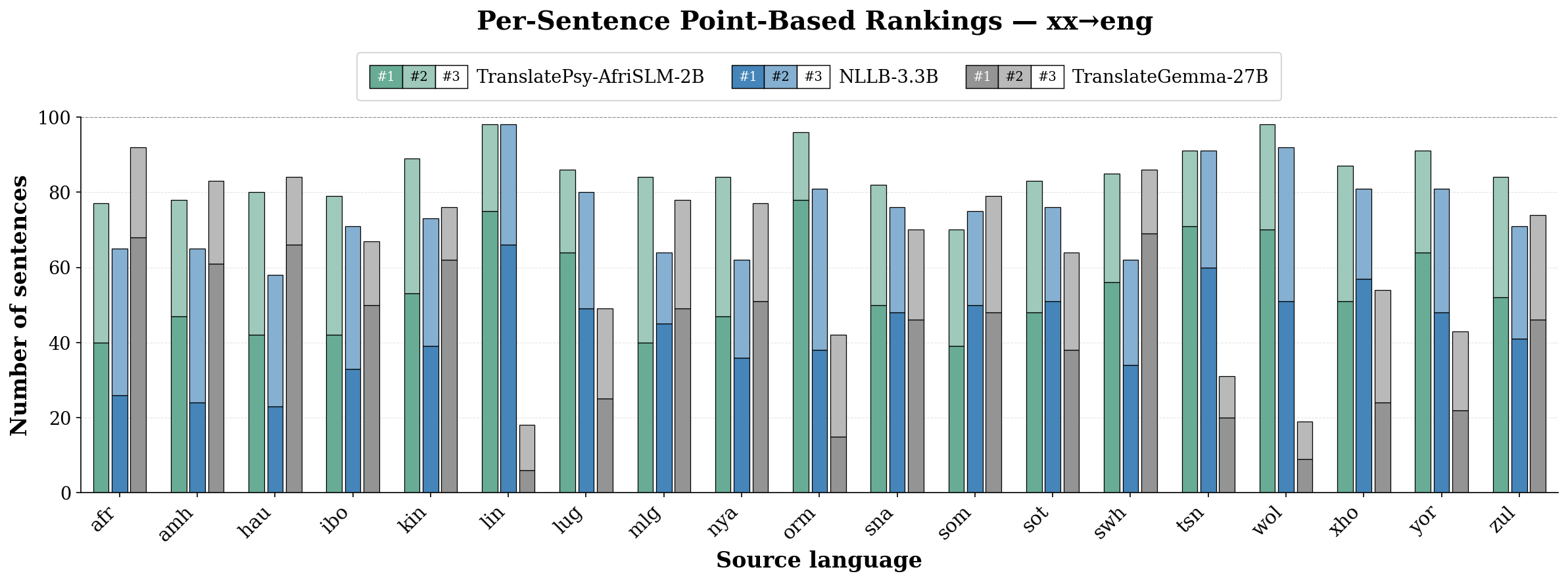}
\vspace{-1.2em}

\caption{\textbf{LLM-as-judge evaluation results.} LLM-as-judge-based rankings for English-to-African language and African language-to-English pairs, respectively, showing the number of times each model was ranked \#1 (darker shade), \#2 (lighter), or \#3 (gap up to the dotted line).}
\label{fig:llm_as_a_judge}
\end{figure*}

Sentence-level rankings among \modelname-2B, NLLB-3.3B, and TranslateGemma-27B were determined using a point system based on evaluator agreement. For each pairwise comparison, a model received 2 points for a unanimous win (both judges preferred the model), 1 point for a split decision (judges disagreed), and 0 points for a loss. Consequently, each model could earn up to 4 points per sentence, allowing for potential ties. Figure~\ref{fig:llm_as_a_judge} presents the language-level breakdown of these LLM-as-a-judge rankings.

\begin{figure*}[t]
\centering
\includegraphics[width=\textwidth]{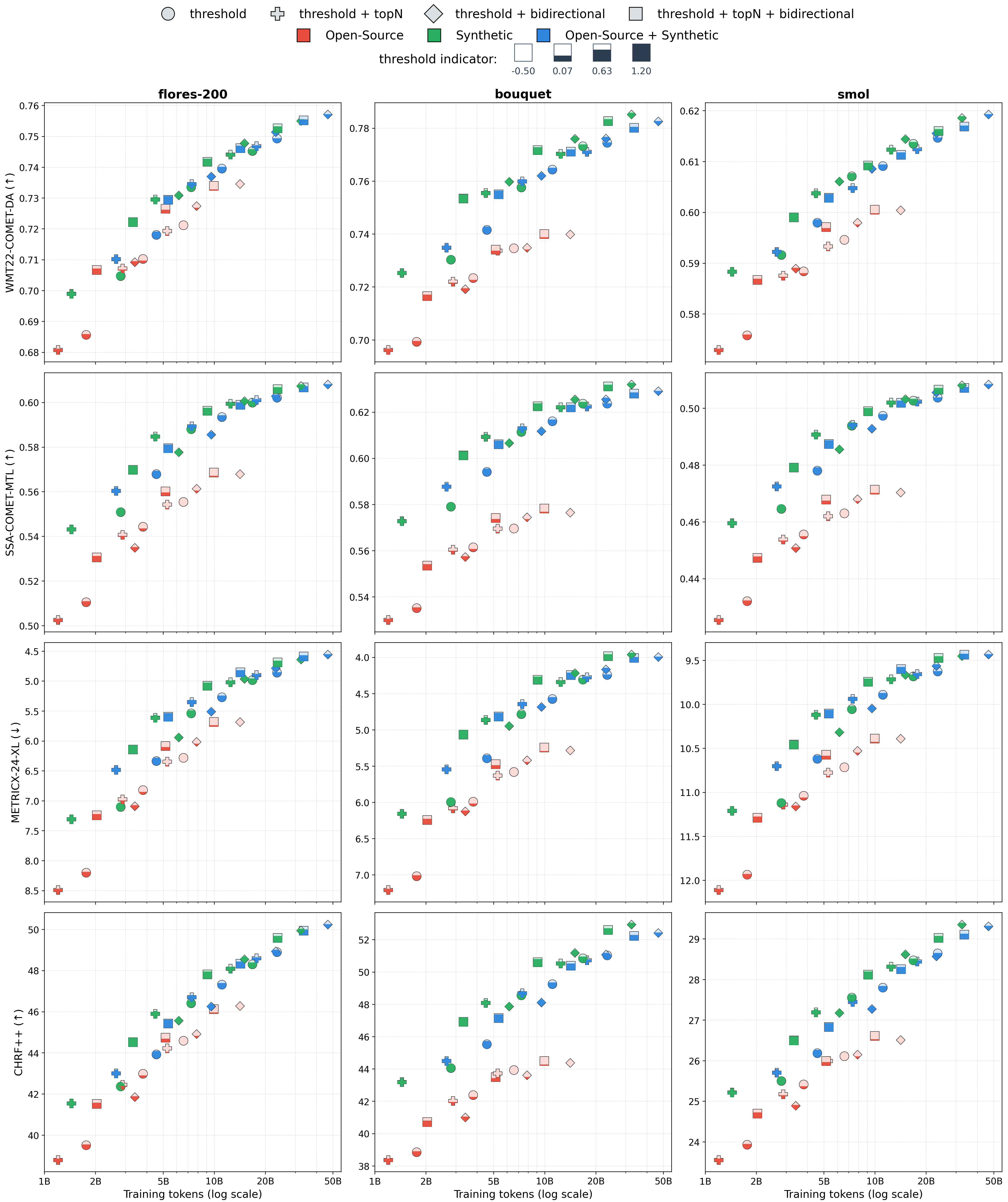}
\caption{\textbf{Scaling runs as QE quality and training budgets increase}, evaluated on Flores-200, BOUQuET, and Smol (columns). Each row reports a different metric, each point corresponds to a model trained with a particular data source and filtering configuration. Colours distinguish filtered, synthetic, and filtered+synthetic data, while marker shapes indicate the filtering setup. Filtering thresholds are represented by an increasing fill, i.e. the fuller the shape, the higher the threshold.}
\label{fig:scaling_eval_all}
\end{figure*}

\begin{table*}[h]
\centering
\begin{tabular}{lcccccc}
\toprule
\multirow{2}{*}{\textbf{Model}} & \multicolumn{2}{c}{\textbf{Flores-200}} & \multicolumn{2}{c}{\textbf{BOUQuET}} & \multicolumn{2}{c}{\textbf{Smol}} \\
\cmidrule(lr){2-3} \cmidrule(lr){4-5} \cmidrule(lr){6-7}
 & SSA (↑) & MX (↓) & SSA (↑) & MX (↓) & SSA (↑) & MX (↓) \\
\midrule
\multicolumn{7}{c}{\textit{\textbf{General-purpose LLMs/SLMs}}} \\
\midrule
Apertus-8B & 0.3814 & 11.471 & 0.4024 & 10.385 & 0.3246 & 14.233 \\
\midrule
Qwen3-4B & 0.2426 & 15.427 & 0.2565 & 14.772 & 0.2117 & 17.273 \\
Qwen3-8B & 0.2655 & 15.327 & 0.2760 & 14.725 & 0.2275 & 17.336 \\
\midrule
Qwen3.5-0.8B & 0.2967 & 15.974 & 0.3320 & 14.902 & 0.2454 & 17.595 \\
Qwen3.5-2B & 0.2885 & 14.344 & 0.3049 & 13.317 & 0.2498 & 15.926 \\
Qwen3.5-4B & 0.3863 & 11.416 & 0.3984 & 10.567 & 0.3301 & 13.868 \\
Qwen3.5-9B & 0.4523 & 9.758 & 0.4664 & 8.927 & 0.3811 & 13.216 \\
Qwen3.5-27B & 0.5132 & 7.250 & 0.5336 & 6.468 & 0.4279 & 11.420 \\
Qwen3.5-122B-A10B & 0.5505 & 5.920 & 0.5716 & 5.318 & 0.4574 & 10.515 \\
\midrule
\multicolumn{7}{c}{\textit{\textbf{Dedicated translation models}}} \\
\midrule
AfriNLLB-600M & 0.5729 & 5.916 & 0.6038 & 4.831 & 0.4748 & 10.476 \\
\midrule
NLLB-600M & 0.5718 & 5.752 & 0.6019 & 4.803 & 0.4723 & 10.430 \\
NLLB-1.3B & 0.5878 & 5.159 & 0.6130 & 4.448 & 0.4850 & 10.036 \\
NLLB-3.3B & 0.5944 & 4.913 & 0.6178 & 4.264 & 0.4909 & 9.841 \\
\midrule
Hunyuan-MT-7B & 0.4450 & 10.779 & 0.4451 & 10.659 & 0.3784 & 13.949 \\
\midrule
TranslateGemma-4B & 0.3913 & 11.086 & 0.4069 & 10.219 & 0.3384 & 14.000 \\
TranslateGemma-12B & 0.5037 & 7.718 & 0.5228 & 6.926 & 0.4299 & 11.501 \\
TranslateGemma-27B & 0.5455 & 6.142 & 0.5677 & 5.506 & 0.4608 & 10.468 \\
\midrule
\multicolumn{7}{c}{\textit{\textbf{African language-specialized models}}} \\
\midrule
%AfriqueLlama-8B & 0.5210 & 7.057 & 0.5620 & 5.679 & 0.4423 & 10.974 \\
%AfriqueGemma-4B & 0.5276 & 6.692 & 0.5587 & 5.716 & 0.4410 & 10.938 \\
%AfriqueGemma-12B & 0.5531 & 5.862 & 0.5805 & 5.093 & 0.4608 & 10.397 \\
%AfriqueQwen-8B & 0.5222 & 6.871 & 0.5549 & 5.748 & 0.4394 & 10.959 \\
%AfriqueQwen-14B & 0.5443 & 6.243 & 0.5718 & 5.408 & 0.4527 & 10.748 \\
AfriqueLlama-8B  & 0.5513 & 5.921 & 0.5772 & 5.181 & 0.4555 & 10.519 \\
AfriqueGemma-4B  & 0.5462 & 6.091 & 0.5729 & 5.313 & 0.4522 & 10.593 \\
AfriqueGemma-12B & 0.5655 & 5.486 & 0.5892 & 4.852 & 0.4649 & 10.286 \\
AfriqueQwen-8B   & 0.5448 & 6.109 & 0.5689 & 5.386 & 0.4503 & 10.655 \\
AfriqueQwen-14B  & 0.5569 & 5.702 & 0.5817 & 5.025 & 0.4605 & 10.348 \\
\midrule
\multicolumn{7}{c}{\textit{\textbf{Ours}}} \\
\midrule
\modelname-0.8B & 0.5944 & 5.197 & 0.6223 & 4.316 & 0.4973 & 9.823 \\
\modelname-2B & 0.6070 & 4.637 & 0.6322 & 3.940 & 0.5074 & 9.458 \\
\modelname-4B & \textbf{0.6143} & \textbf{4.317} & \textbf{0.6391} & \textbf{3.701} & \textbf{0.5136} & \textbf{9.232} \\
\bottomrule
\end{tabular}
\vspace{0.5em}
\caption{\textbf{SSA-COMET and MetricX} results on Flores-200, BOUQuET, and Smol.}
\label{tab:ssa_results}
\end{table*}

\begin{table*}[t]
\centering
\small
\resizebox{\textwidth}{!}{
\begin{tabular}{lccccccccc}
\toprule
\multirow{2}{*}{\textbf{Model}} & \multicolumn{3}{c}{\textbf{Flores-200}} & \multicolumn{3}{c}{\textbf{BOUQuET}} & \multicolumn{3}{c}{\textbf{Smol}} \\
\cmidrule(lr){2-4} \cmidrule(lr){5-7} \cmidrule(lr){8-10}
 & C22 (↑) & chrF++ (↑) & spBLEU (↑) & C22 (↑) & chrF++ (↑) & spBLEU (↑) & C22 (↑) & chrF++ (↑) & spBLEU (↑) \\
\midrule
\multicolumn{10}{c}{\textit{\textbf{General-purpose LLMs}}} \\
\midrule
Apertus-8B & 0.5883 & 30.45 & 12.22 & 0.6025 & 27.81 & 11.06 & 0.4912 & 18.01 & 3.52 \\
\midrule
Qwen3-4B & 0.4046 & 17.35 & 5.03 & 0.4305 & 14.24 & 3.59 & 0.3607 & 10.74 & 0.72 \\
Qwen3-8B & 0.4566 & 20.63 & 6.90 & 0.4792 & 17.29 & 5.19 & 0.4002 & 12.49 & 1.17 \\
\midrule
Qwen3.5-0.8B & 0.3656 & 14.06 & 3.27 & 0.3920 & 11.16 & 2.12 & 0.3255 & 8.61 & 0.38 \\
Qwen3.5-2B & 0.4724 & 20.95 & 5.87 & 0.4934 & 17.82 & 4.54 & 0.4143 & 12.90 & 1.05 \\
Qwen3.5-4B & 0.6021 & 30.58 & 11.35 & 0.6122 & 27.88 & 10.28 & 0.5103 & 18.58 & 3.09 \\
Qwen3.5-9B & 0.6688 & 36.74 & 15.52 & 0.6801 & 34.96 & 15.24 & 0.5566 & 22.19 & 4.98 \\
Qwen3.5-27B & 0.7144 & 41.78 & 19.96 & 0.7264 & 41.22 & 20.52 & 0.5859 & 24.93 & 7.33 \\
Qwen3.5-122B-A10B & 0.7373 & 44.42 & 21.89 & 0.7494 & 44.29 & 22.98 & 0.6045 & 26.67 & 8.46 \\
\midrule
\multicolumn{10}{c}{\textit{\textbf{Dedicated translation models}}} \\
\midrule
AfriNLLB-600M & 0.7315 & 46.74 & 25.76 & 0.7671 & 48.82 & 29.89 & 0.6008 & 27.99 & 11.07 \\
\midrule
NLLB-600M & 0.7340 & 45.73 & 24.48 & 0.7654 & 47.83 & 28.44 & 0.6028 & 27.08 & 9.85 \\
NLLB-1.3B & 0.7471 & 47.77 & 26.80 & 0.7753 & 49.66 & 30.66 & 0.6101 & 27.98 & 10.94 \\
NLLB-3.3B & 0.7525 & 48.89 & 28.06 & 0.7795 & 52.16 & 33.75 & 0.6139 & 28.60 & 11.50 \\
\midrule
Hunyuan-MT-7B & 0.5056 & 23.49 & 6.03 & 0.5211 & 19.56 & 4.53 & 0.4391 & 15.49 & 1.19 \\
\midrule
TranslateGemma-4B & 0.5487 & 24.78 & 10.21 & 0.5699 & 23.12 & 10.65 & 0.4644 & 15.21 & 3.02 \\
TranslateGemma-12B & 0.7125 & 39.48 & 16.66 & 0.7211 & 38.91 & 18.03 & 0.5983 & 24.56 & 5.97 \\
TranslateGemma-27B & 0.7335 & 43.33 & 21.13 & 0.7499 & 43.72 & 23.49 & 0.6102 & 26.24 & 8.12 \\
\midrule
\multicolumn{10}{c}{\textit{\textbf{African language-specialized models}}} \\
\midrule
%AfriqueLlama-8B & 0.6689 & 37.58 & 15.04 & 0.7179 & 41.58 & 22.35 & 0.5630 & 23.02 & 6.72 \\
%AfriqueGemma-4B & 0.6916 & 40.75 & 21.51 & 0.7223 & 42.51 & 24.98 & 0.5688 & 23.56 & 7.75 \\
%AfriqueGemma-12B & 0.7131 & 44.53 & 25.20 & 0.7429 & 46.22 & 28.65 & 0.5847 & 25.96 & 9.98 \\
%AfriqueQwen-8B & 0.6754 & 38.23 & 18.03 & 0.7168 & 41.09 & 23.21 & 0.5659 & 23.06 & 6.84 \\
%AfriqueQwen-14B & 0.6955 & 41.58 & 21.44 & 0.7297 & 43.67 & 26.14 & 0.5754 & 24.64 & 8.50 \\
AfriqueLlama-8B & 0.7245 & 45.04 & 25.57 & 0.7513 & 45.90 & 28.97 & 0.5913 & 25.71 & 9.82 \\
AfriqueGemma-4B & 0.7203 & 44.46 & 24.98 & 0.7466 & 45.21 & 28.14 & 0.5896 & 25.44 & 9.49 \\
AfriqueGemma-12B & 0.7389 & 47.84 & 28.48 & 0.7636 & 48.46 & 31.36 & 0.5999 & 27.15 & 11.22 \\
AfriqueQwen-8B & 0.7202 & 44.41 & 25.01 & 0.7448 & 45.20 & 28.20 & 0.5878 & 25.43 & 9.54 \\
AfriqueQwen-14B & 0.7303 & 46.16 & 26.82 & 0.7553 & 46.96 & 29.91 & 0.5968 & 26.36 & 10.28 \\
\midrule
\multicolumn{10}{c}{\textit{\textbf{Ours}}} \\
\midrule
\modelname-0.8B & 0.7432 & 48.30 & 27.00 & 0.7770 & 51.30 & 32.35 & 0.6120 & 28.54 & 11.21 \\
\modelname-2B & 0.7552 & 50.00 & 28.95 & 0.7870 & 53.13 & 34.49 & 0.6187 & 29.32 & 12.04 \\
\modelname-4B & \textbf{0.7629} & \textbf{51.13} & \textbf{30.28} & \textbf{0.7941} & \textbf{54.33} & \textbf{36.07} & \textbf{0.6237} & \textbf{29.97} & \textbf{12.80} \\
\bottomrule
\end{tabular}
}
\vspace{0.5em}
\caption{\textbf{COMET-22, chrF++, and spBLEU} results on Flores-200, BOUQuET and Smol.}
\label{tab:chrf_results}
\end{table*}

\begin{figure*}[t]
\centering
\vspace{-3.5mm}
\includegraphics[width=1\textwidth]{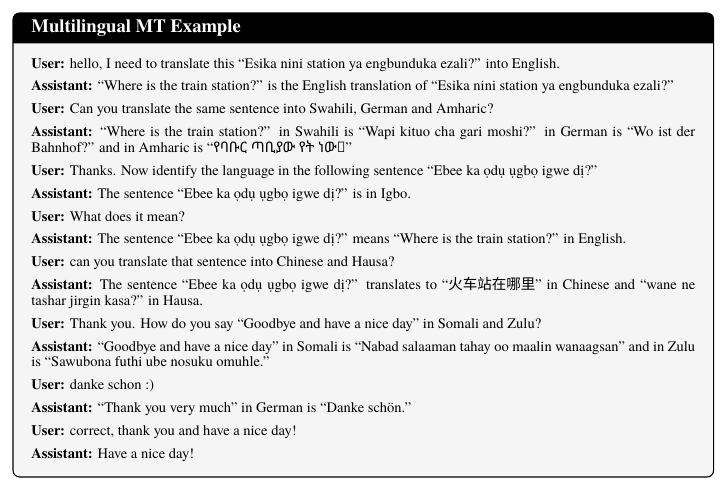}
\vspace{-2mm}
\caption{Multi-turn and cross-lingual translation and language identification with \modelname-4B.}
\label{fig:multilingual_mt_conversation}
\end{figure*}

% REBUTTAL --------------------------------------

\end{document}